\documentclass[10pt,twocolumn,letterpaper]{article}

\usepackage[pagenumbers]{iccv}

\newcommand{\horus}{ARGUS\xspace}

\newcommand{\hscore}{ArgusCost\xspace}
\newcommand{\hscoreh}{ArgusCost-H\xspace}
\newcommand{\hscoreo}{ArgusCost-O\xspace}

\newcommand{\hbench}{ArgusBench\xspace}
\definecolor{iccvblue}{rgb}{0.21,0.49,0.74}
\definecolor{rs}{rgb}{0.2813, 0.1953, 0.2813}
\usepackage[pagebackref,breaklinks,colorlinks,allcolors=iccvblue]{hyperref}

\usepackage{amsmath}

\usepackage{amsmath, amssymb, amsthm}
\usepackage{bm}

\usepackage[utf8]{inputenc}   
\usepackage{booktabs}         
\usepackage{array}            
\usepackage{listings}
\usepackage{pifont}

\usepackage{tcolorbox}
\tcbuselibrary{breakable}

\let\oldding\ding
\renewcommand{\ding}[2][1]{\scalebox{#1}{\oldding{#2}}}
\definecolor{lightroyalblue}{HTML}{F6F8FD} 
\definecolor{royalblue}{HTML}{4169E1}
\definecolor{lighterblue}{HTML}{f2fafd}  
\newtcolorbox{abox}{colback=lightroyalblue,colframe=black}

\definecolor{darkblue}{RGB}{46,25, 110}
\usepackage{soul}

\hypersetup{ 
    colorlinks=true,
    linkcolor=magenta,
    filecolor=magenta,      
    urlcolor=magenta,
    citecolor=teal,
    pdftitle={CinePile A long video question answering dataset and benchmark},
    pdfpagemode=FullScreen,
    }

\def\paperID{9086} 
\def\confName{ICCV}
\def\confYear{2025}

\title{\horus: Hallucination and Omission Evaluation in Video-LLMs}

\author{
Ruchit Rawal \quad
Reza Shirkavand \quad
Heng Huang \quad
{Gowthami Somepalli}$^\star$ \quad 
{Tom Goldstein}$^\star$
\\
\and 
University of Maryland, College Park
\\ \and
\url{https://hf.co/datasets/tomg-group-umd/argus}
}

\begin{document}
\maketitle

\def\thefootnote{}\footnotetext{\vspace{-.1in}\\$^\star$Equal contribution. Correspondence: \url{ruchitr@umd.edu}}
\def\thefootnote{\arabic{footnote}}

\begin{abstract}
Video large language models have not yet been widely deployed, largely due to their tendency to hallucinate.  Typical benchmarks for Video-LLMs rely simply on multiple choice questions. Unfortunately, VideoLLMs hallucinate far more aggressively on freeform text generation tasks like video captioning than they do on multiple choice verification tasks.  To address this weakness, we propose \horus, a VideoLLM benchmark that measures freeform video captioning performance.  
By comparing VideoLLM outputs to human ground truth captions, \horus quantifies dual metrics.  First, we measure the rate of hallucinations in the form of incorrect statements about video content or temporal relationships.  Second, we measure the rate at which the model omits important descriptive details.  Together, these dual metrics form a comprehensive view of video captioning performance.
\end{abstract}
    
\section{Introduction}
\label{sec:intro}

Video Large Language Models (VideoLLMs)~\cite{gpt4o,team2023gemini,internvl2} have made significant strides in recent years, and improvements in their capabilities have been reflected in rising scores on benchmarks for both short video~\cite{perceptiontest,caba2015activitynet} and long video~\cite{wang2024lvbench,cinepile,videomme,egoschema,nagrani2025minerva,nagrani2024neptune}. Despite these strides, VideoLLMs are not yet ready for widespread deployment.
The main challenge is their tendency to hallucinate, with the frequency and severity of these hallucinations increasing along with the size and complexity of the input video. 
 
 \looseness=-1 
Recent studies~\cite{langmodelsbetter, kaul2024throne} indicate that while language models are often effective as verifiers that provide yes/no or multiple choice answers, they are less reliable as freeform generators. This disparity highlights a crucial limitation: a VideoLLM’s ability to verify the presence of an object or the occurrence of an event does not necessarily translate into open-ended tasks like dense video captioning. The latter is particularly important for models that assist users with perceptual disabilities. Hence, there is a pressing need for a dedicated benchmark that evaluates VideoLLMs on their freeform generative capabilities rather than solely on question-answering.
\begin{figure}
    \centering
    \includegraphics[width=\linewidth]{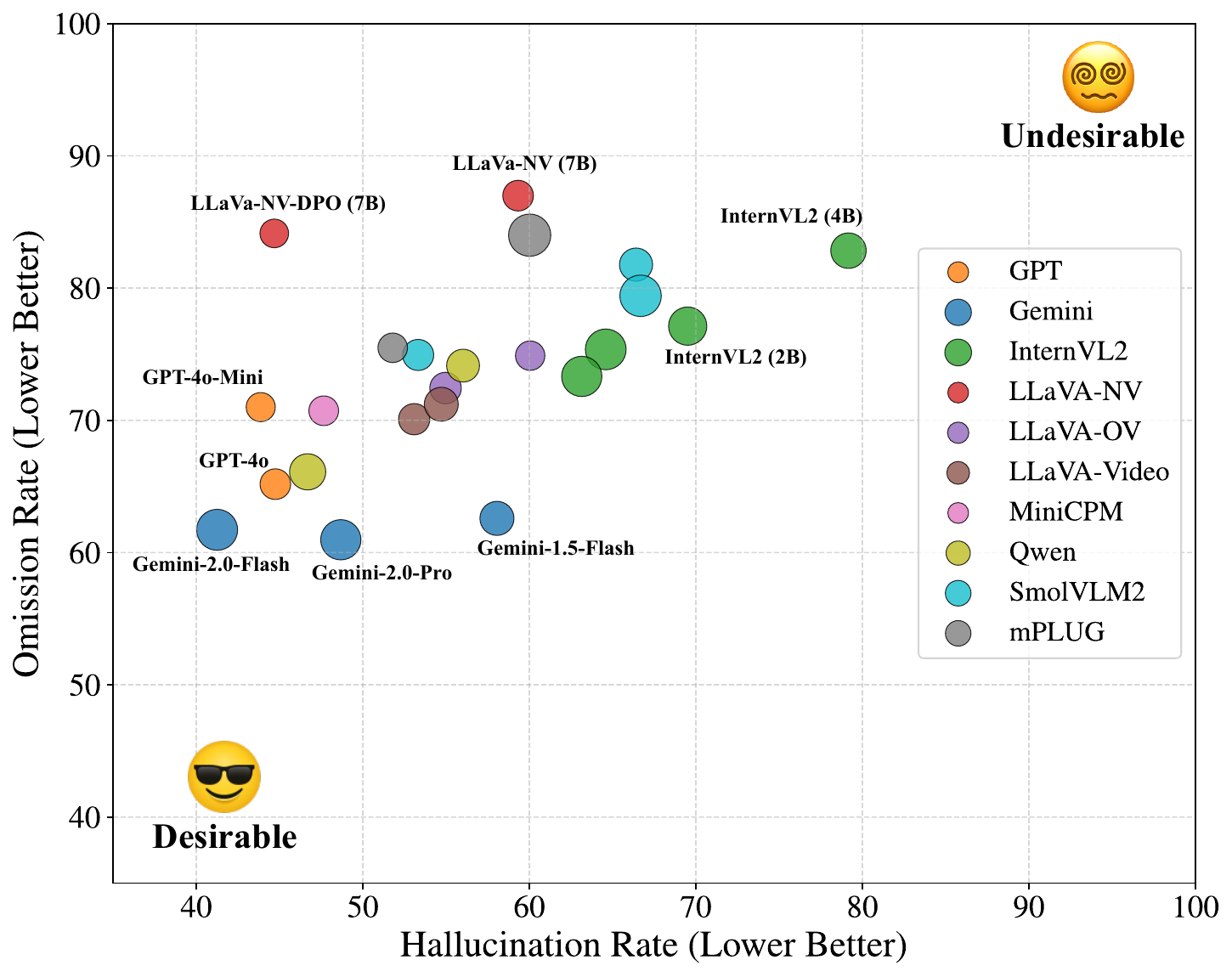}
    \caption{\textbf{Relationship between Hallucination and Omission}. The hallucination and omission cost metrics are correlated; however, most models exhibit more omissions than hallucinations. Marker size indicates the average caption length per model. Gemini-2.0-Flash achieves the best performance.}
    \label{fig:hallucination_v_omission_model_familiy}
    \vspace{-1em}
\end{figure}

In this work, we develop \horus\footnote{Named after Argus, the hundred-eyed giant of Greek mythology renowned for his vigilance and ability to monitor every detail.}, a novel benchmark that evaluates the rate of hallucination in free-form video captions from VideoLLMs. Unfortunately, measuring hallucination alone is problematic --- a model can avoid falsehoods simply by generating the empty string, and so it is meaningless to measure freeform accuracy without a dual metric of completeness. 
Hallucinations and omissions represent two sides of the same fundamental challenge — ensuring both the accuracy and completeness of video understanding.
To the best of our knowledge, no benchmarks currently exist that systematically evaluate hallucinations and omissions in a freeform setting for VideoLLMs. 

Specifically, to address this gap, the \horus framework compares the sentences from the VideoLLM generated caption to the human sentences, and an entailment analysis is used to quantify the rate hallucinations in the form of (i) inaccurate summarization, (ii) incorrect visual descriptions, and (iii) incorrect depiction of temporal relationships.  At the same time, the VideoLLM caption is analyzed to ensure that it contains all the important statements extracted from the human caption, and the rate of omissions is quantified. 

Our goal with \horus is to provide a fair platform for benchmarking and comparing models in a freeform setting, as depicted in~\cref{fig:hallucination_v_omission_model_familiy}.
We posit that enhancing the accuracy and comprehensiveness of dense captioning will naturally bolster other downstream tasks such as VideoQA and reasoning, as a more holistic understanding of video content is inherently acquired. Dataset and artifacts are available at \url{https://ruchitrawal.github.io/argus}.
\begin{figure}[t!]
    \centering
    \includegraphics[width=\linewidth]{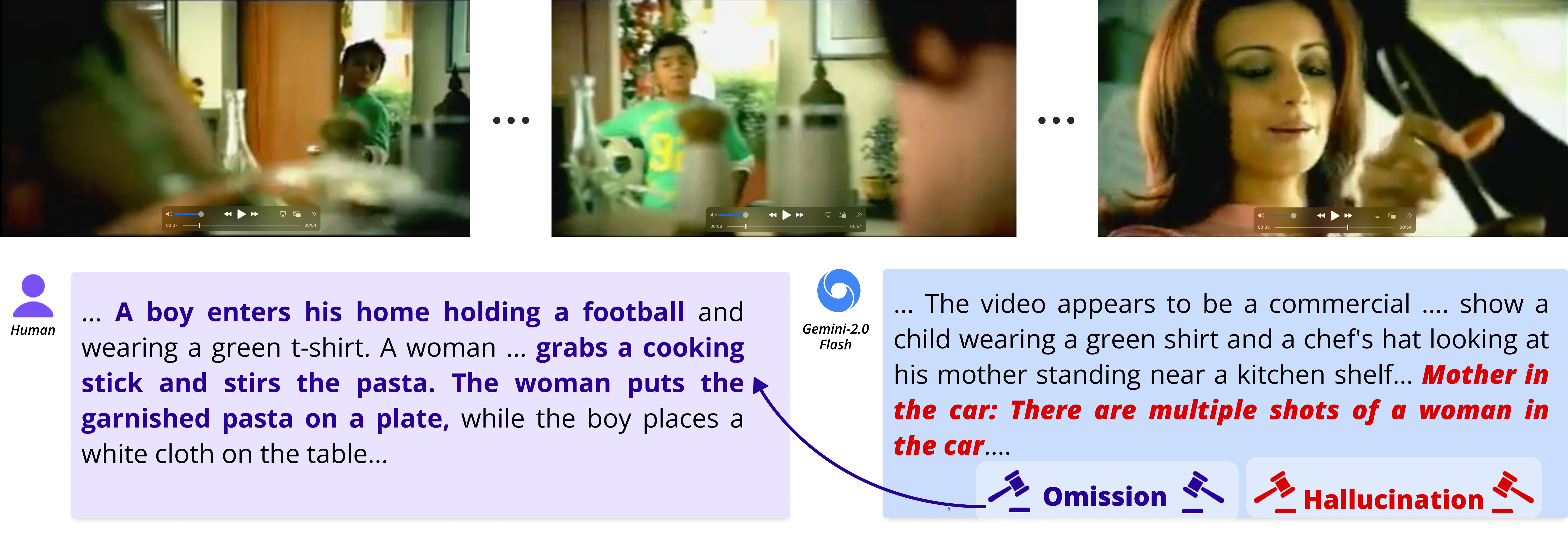}
    \caption{An example annotation (see \href{https://www.youtube.com/watch?v=uOk4EFDsDP4}{here}) of a video by the Gemini-2-Flash model. Using our framework \horus, we identify both hallucinations and omissions in this dense caption. We present the full human and model-captions in Appendix~\ref{appendix_sec:details_qual_examples}.}
    \label{fig:omi-hal-qual}
    \vspace{-10pt}
\end{figure}

\section{Related Work}
\label{sec:related_work}

\looseness=-1 
\textbf{Hallucination Evaluations in Vision.} Many benchmarks such as CHAIR~\cite{rohrbach2018object} and others~\cite{wang2023amber,jing2023faithscore,hu2023ciem} are object-centric, employing heuristics to determine the presence or absence of specific objects. Other approaches, like POPE~\cite{li2023evaluating}, use Yes/No questions to assess how well a model understands a video, while others~\cite{sun2023aligning,liu2023mitigating} have integrated large language models into their pipelines for end-to-end annotations. These strategies often encounter challenges related to scalability, incomplete evaluations, or the introduction of additional errors stemming from the limitations of using LLMs as end-to-end evaluators. For a more comprehensive discussion see~\cite{liu2024survey}.

 \looseness=-1 
While understanding hallucinations is a bit mature research topic in Image captioning models, doing the same for videos is still very new. VideoHallucer~\cite{wang2024videohallucer}, one of the first works, evaluates hallucinations in Video-LLMs using an adversarial binary question-answering approach, where a model is considered to be hallucinating if it answers either a basic or a carefully designed hallucinated question incorrectly. EventHallusion~\cite{guan2024hallusionbench} follows a similar question-answering-based strategy but extends it to both binary and open-ended questions, focusing primarily on short, single-event videos averaging 11 seconds in length. VIDHAL~\cite{choong2024vidhal} takes a different approach, proposing a caption reordering task to assess whether Video-LLMs can verify whether a caption contains more hallucinated content than another.

 \looseness=-1 
\noindent Existing benchmarks measure a model's ability to verify the presence of content, we focus on the more challenging task of generating open-ended captions. Additionally, unlike prior works, we measure both hallucination and omission. 
We compare our benchmark with other works in~\cref{tab:video-benchmarks}, demonstrating that we are the only dedicated benchmark for dense captioning. Moreover, we do not lose out on the number of videos or video length while providing more fine-grained evaluations than other benchmarks.

\begin{table}[ht]
\centering
\scalebox{0.9}{  
\begin{tabular}{p{3.0cm} p{0.9cm} p{0.9cm} p{1.7cm} p{0.7cm}}
\toprule
\textbf{Dataset} & \textbf{\# Vid.} & \textbf{Avg. Sec} & \textbf{Eval. Strategy} & \textbf{Evals \/ Video} \\
\midrule
HallusionBench \cite{guan2024hallusionbench} & 20 & $\leq$4 & QA & 8 \\
VideoHallucer \cite{wang2024videohallucer} & 948 & 85.6 & QA & 1.89 \\
VIDHAL \cite{choong2024vidhal} & 1000 & 15.8 & Capt. Order & 1 \\
Event-Hallu. \cite{zhang2024eventhallusion} & 397 & 11.2 & QA & 1.77 \\
\midrule
\textbf{\horus (Ours)} & 500 & 26.3 & Dense-Cap &  19 \\
\bottomrule
\end{tabular}
}
\caption{\textbf{Comparison of various video-hallucination benchmarks.} ``Evals Video" refers to the number of evaluations done on average per video, task could be different depending on the benchmark. Refer to \cref{sec:related_work} for the exact tasks.}
\label{tab:video-benchmarks}
\vspace{-0.5em}
\end{table}

\begin{figure*}[t!]
    \centering
    \includegraphics[width=\linewidth]{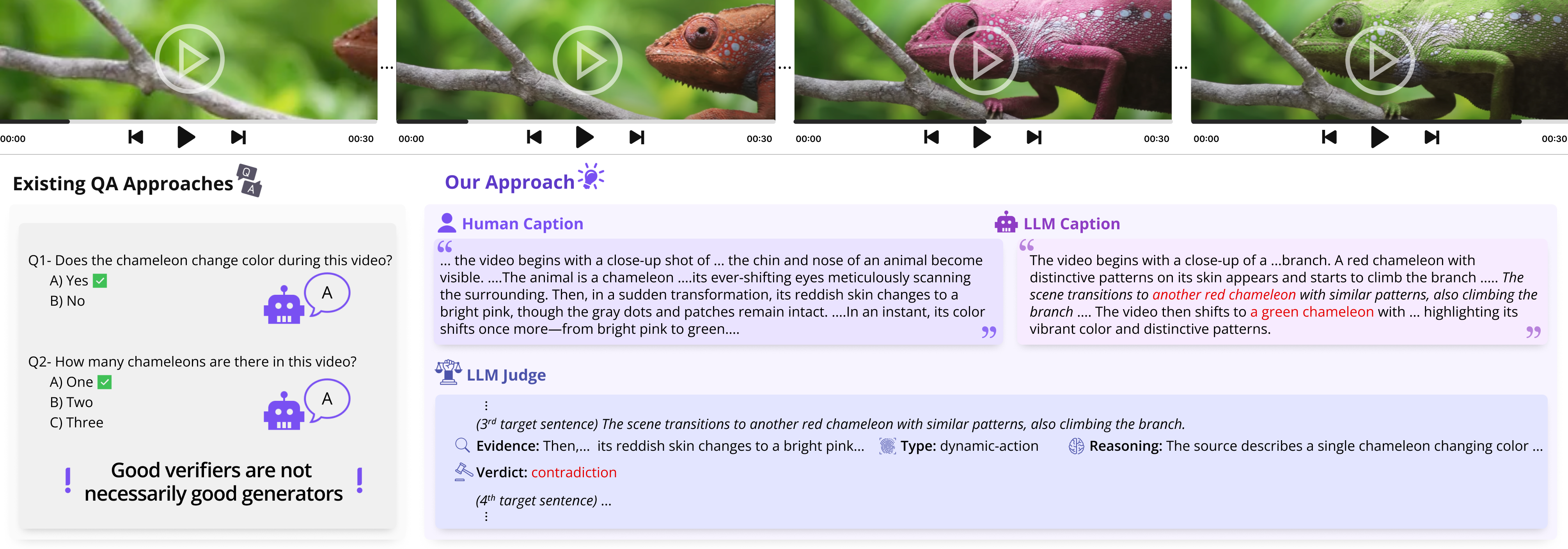}
    \caption{A Video LLM can correctly answer targeted questions about a video (left; see \href{https://www.youtube.com/watch?v=ioblgpA5eTo}{here}) but still generate hallucinated content when describing the video (right), highlighting the disconnect between verification and open-ended generation. Our approach evaluates hallucinations at the sentence level from the generated caption, providing a more precise and comprehensive assessment. Refer to \cref{subsec:sent_level_entailment} to see how we ground each sentence using an LLM judge to compute a hallucination score. Example LLaVa-OV-7B caption and response on a video from \hbench. We present the full human and model-captions in Appendix~\ref{appendix_sec:details_qual_examples}} 
    \label{fig:teaser_figure}
    \vspace{-1em}
\end{figure*}

\noindent  \textbf{Limitations of Current QA Approaches.} 
Existing benchmarks rely on a question-answering (QA) paradigm, where the model is tested with two types of questions: one basic type that expects a ``Yes" (confirming the presence of entities) and another  that expects a ``No" (flagging potential hallucinations). Although this setup simplifies evaluation, it suffers from the following limitations - \textbf{(1) Lack of Dependence on Visual Understanding:} Many binary QA-based evaluations~\cite{lin2023revisiting, zohar2024apollo} contain questions that can be answered without processing the visual input. To illustrate this, in our experiment with GPT-4o on a subset of VideoHalluCer \cite{wang2024videohallucer}, the text-only model correctly answered 32.52\% of basic hallucination-related question pairs~(with 61.33\% accuracy in the \texttt{external\_nonfactual\_instruct} subcategory), despite a 25\% chance performance. \textbf{(2) Verification Ability Does Not Equate to Strong Generation:} Video LLMs can verify facts in a QA setting yet fail in open-ended generation; for instance, while LLaVa-OV-7B~\cite{llavaonevision} verified that there is one chameleon in a clip, it mistakenly generated a caption describing two \cite{kaul2024throne} (see~\cref{fig:teaser_figure} for a qualitative example and Appendix~\ref{appendix_sec:method_sub:current-approaches} for quantitative results). \textbf{(3) Restricted Error Coverage Due to Predefined Scope:} The QA-based approach is limited by its predefined set of questions, covering only a narrow range of errors and leaving many hallucinations undetected until free-form captions are generated. \textbf{(4) Inability to Capture Multi-Event Hallucinations:} QA-based methods, often focusing on isolated short events, do not account for complex interactions and temporal dependencies in videos with multiple interrelated events (e.g., four sequential events with a 50\% chance of guessing correctly in a binary setting), unlike free-form generation strategies.
Please refer to~\cref{appendix_sec:method_sub:current-approaches} for expanded discussion on this topic and~\cref{appendix_sec:extended_related_work} for related work on Natural Language Inference and Dense Video Captioning.

\section{\horus: Freeform Captioning Benchmark}
\label{sec:methods}
We propose \horus, a novel evaluation framework to quantify both hallucinations and omissions in VideoLLM captions that addresses the drawbacks of previous benchmarks. We introduce a metric \hscoreh that quantifies a model's average rate of hallucination, and a metric \hscoreo that reflects the rate of omissions in VideoLLM. In the remainder of this section, we detail the computation of the hallucination metric, \hscoreh.
 We use a similar methodology to compute \hscoreo.

\looseness=-1 
We analyze VideoLLMs captions by matching selected sentences to a corresponding sentence in a ground-truth human caption. Using this matching, we identify two types of hallucinations.  First, we consider hallucinations where the model incorrectly recognizes {\em false content} in the video.  We identify such errors by using Natural Language Entailment (with an LLM as a judge) between the VideoLLM sentence and its matched ground truth. Second, we penalize hallucinations in which the sentences are correct on their own, but the model misrepresents the {\em order of events}. It is important to score these errors, as time is the key difference between image and video understanding.  We check whether the order of sentences that make temporal claims in the VideoLLM caption matches the order of their corresponding sentences in the human caption, and assign a penalty to each sentence that is proportional to its level of anachronism. 

\looseness=-1 
If the model fabricates an event that did not happen, this should be considered a content error and not a temporal error, and therefore left out of the temporal matching. However, if such a fabrication gets mistakenly matched to a similar but faraway sentences in the ground truth, this will disrupt the matching between otherwise correct temporal statements and cause artificially large error scores.  To prevent hallucinated events from disrupting the temporal matching, we use a dynamic program that assigns each event to either the content or temporal error category in order to minimize the overall temporal matching penalty. See~\cref{subsec:sent_level_entailment,subsec:dp_formulation}.

\subsection{Sentence-Level Entailment}
\label{subsec:sent_level_entailment}
 \looseness=-1 
For a given video, we assume access to a set of $m$ source sentences annotated by human annotators, i.e., $S = \{s_1, s_2, \dots, s_m\}$, describing the video in detail. Similarly, we also assume access to $n$ target sentences generated by a VideoLLM, i.e., $T = \{t_1, t_2, \dots, t_n\}$. Our goal is to determine whether any of these target sentences contain hallucinated content that has no grounding in $S$. If a target sentence $t_i$ is entailed by the source set $S$, then it is considered valid; otherwise, it is hallucinated. 
LLMs and reasoning models now match human performance on complex logical reasoning tasks \cite{liu2025logical, guo2025deepseek} and are increasingly used as ``judges" in dynamic evaluation pipelines \cite{li2024llms, zheng2023judging}. We, hence use a strong model, GPT-4o, as the entailment judge. We tested alternatives like DeepSeek-V3 and the reasoner DeepSeek-R1 in~\cref{subsec:sensitivity_analysis}, finding similar results. 

 \looseness=-1 
\noindent We input both $S$ and $T$ to an LLM-judge, which evaluates each target sentence $t_i \in T$ along three key dimensions: 
\begin{itemize}
   \item A type \(\theta_i \in \{\text{SUM}, \text{VD}, \text{DA}\}\), categorizing \(t_i\) as either a summary (SUM), a visual description (VD), or a dynamic action (DA).
    \item A verdict \(v_i \in \{\text{EN}, \text{CON}, \text{UD}\}\), indicating whether \(t_i\) is entailed (EN), contradictory (CON), or undetermined (UD) with respect to \(S\).
    \item An evidence line \(e_i \in \{1,2,\dots,m\} \cup \{\text{null}\}\), corresponding to the location of a supporting sentence in \(S\) or marked as null if no supporting evidence exists.
\end{itemize}

\vspace{-1.2em}
 \looseness=-1 
\paragraph{Why do we need type and verdict categorization?} \horus scores sentences differently depending on their type. Dynamic actions have an inherent temporal structure, requiring a penalty when their order described in the target caption deviates from the human caption. In contrast, summarization and visual-descriptions do not follow a strict temporal order, but should still be checked for entailment. 

\vspace{-0.2em}
\subsection{Calculating Total Cost}
\label{subsec:dp_formulation}
We first separately define for two key components of the hallucination cost: (1) the base cost, which captures the penalty for sentences that violate entailment, and (2) the order penalty, which accounts for temporal misalignment in dynamic actions. We then unify these costs using a dynamic programming formulation that computes the optimal alignment between the source caption $S$ and the generated caption $T$. 

\subsubsection{Base Cost Matrix}  
We define a cost matrix \( C \in \mathbb{R}^{n \times m} \), where each entry represents the hallucination cost for a target sentence \( t_i \) given its evidence in the source sentence \( s_j \):  

\vspace{-1.5em}
\begin{equation}
\resizebox{0.9\linewidth}{!}{$
C_{ij} =
\begin{cases}
  1, & \text{if } v_i \neq \text{EN}, \\[1mm]
  0, & \text{if } v_i = \text{EN} \text{ and } \theta_i \in \{\text{SUM},\ \text{VD}\}, \\[1mm]
  \begin{cases}
     0, & \text{if } j = e_i, \\
     1, & \text{otherwise},
  \end{cases} & \text{if } v_i = \text{EN} \text{ and } \theta_i = \text{DA}.
\end{cases}
$}
\end{equation}

\begin{figure}
    \centering
    \includegraphics[width=\linewidth]{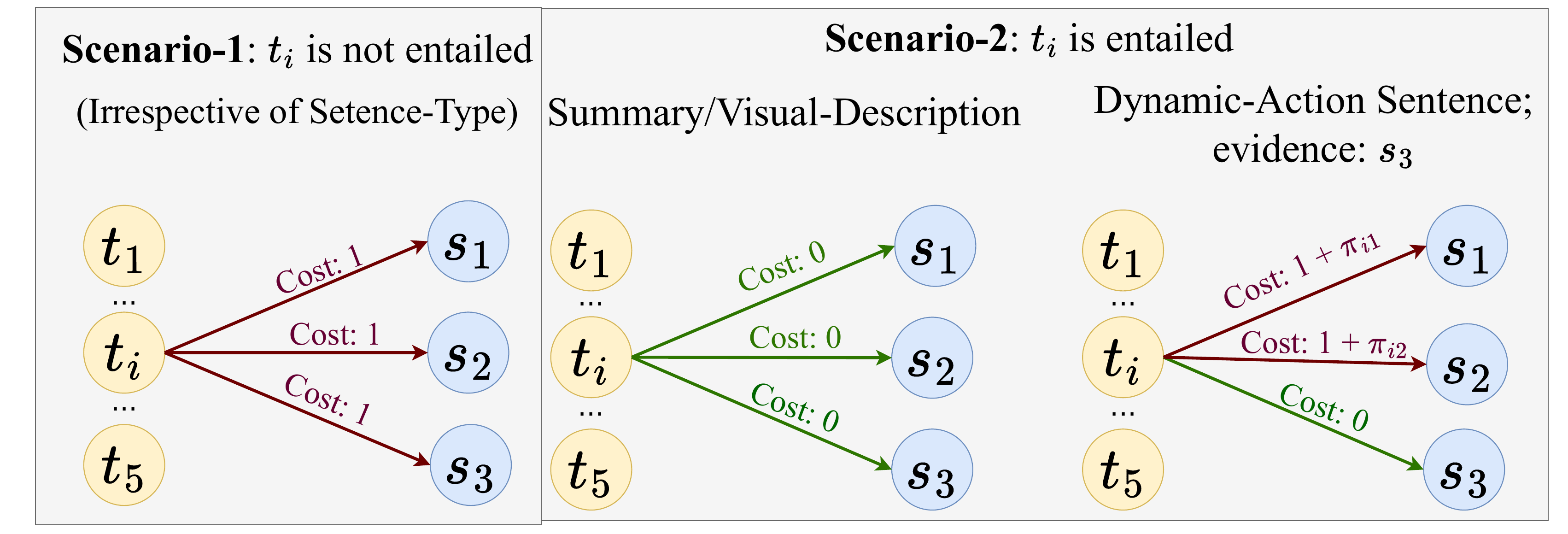}
    \caption{\textbf{Proposed dynamic programming formulation for sentence alignment.} Given sentence-level entailment scores from the LLM-as-a-judge, we compute the matching cost between a target sentence ($t_i$) and a source sentence ($s_j$) by combining the entailment score with a temporal order penalty ($\pi_{ij}$). The overall score follows the recurrence relationship defined in~\cref{subsec:dp_formulation}.}
    \label{fig:dp_method_figure}
    \vspace{-7pt}
\end{figure}

Visualized in~\cref{fig:dp_method_figure}, the Base Cost Matrix primarily captures the cost associated with hallucinations based on entailment labels. If $t_i$ is not entailed ($v_i \neq \text{EN}$), we assign a maximum penalty of $1$. Conversely, if $t_i$ is entailed and belongs to the summary (SUM) or visual description (VD) category, we assign a cost of 0. This is because summary sentences often do not correspond to a specific source sentence but rather capture the overall essence of the video, while visual descriptions provide general scene details that do not adhere to a strict temporal structure. 

For dynamic actions (DA), however, temporal alignment matters. If $t_i$ is entailed, we check whether its supporting evidence $e_i$ is the same as the source index $j$. If $j = e_i$, we assign a cost of 0, indicating a correct match. Otherwise, we impose a penalty of $1$, reflecting a misalignment in the order in which events are described.

\subsubsection{Dynamic Programming Formulation}

\textbf{DP State:} We define a DP table $D \in \mathbb{R}^{(n+1) \times m}$ where each entry $D(i,j)$ represents the minimum cost of aligning the first $i$ target sentences with the source sentences, such that the $i$-th target is matched with source sentence $s_j$. In addition to tracking the cost, we also need to maintain information about the matching structure—specifically, which source sentences were matched with previous target sentences. This alignment history is necessary for computing ordering penalties accurately.

\noindent \textbf{Alignment History:} Let $A_{i,j}$ denote the sequence of source indices $(a_1, a_2, \ldots, a_{i-1}, j)$ that represents the optimal alignment of the first $i$ target sentences, with the $i$-th target aligned to source $s_j$. This alignment history enables us to evaluate temporal consistency in the sequence of described actions.

\noindent \textbf{Ordering Penalty:} The ordering penalty function quantifies temporal inconsistencies in the alignment. Given an alignment history $A_{i,k}$, we define:
\begin{equation}
    \begin{split}
        \pi(A_{i,k}, j) =
\lambda \cdot \sum_{r \leq i} \mathbf{1}\Bigl[
a_r > j \Bigr] \\ \text{s.t. } \theta_r = \theta_j = \text{DA}
\quad \text{and } v_r = v_j = \text{EN},
    \end{split}
\end{equation}

where $0\leq \lambda \leq1$ is a penalty factor, and $\mathbf{1}[\cdot]$ is the indicator function. This function counts ordering violations between pairs of dynamic actions. A violation occurs when an earlier action in the target sequence is aligned with a later action in the source sequence, relative to another action pair. The severity of these violations increases with the number of pairs that appear out of order, making sequences with multiple ordering inversions more heavily penalized than those with fewer inversions. 

Intuitively, as $\lambda$ increases, \horus becomes less tolerant of temporal inconsistencies; sentences placed significantly out of their correct order may effectively be treated as hallucinations during the DP-recurrence minimization. In our setup, we use $\lambda = 0.1$. 

\noindent \textbf{Recurrence Relation:} 
Intuitively, $D(i,j)$ equals the total cost of aligning the previous $i-1$ sentences ($D(i-1, .)$) plus the base cost of aligning the $i^{th}$ with the $j^{th}$ source sentence, and an ordering penalty if misalignment occurs. Formally, the recurrence relation for each $i \in {1, 2, \ldots, n}$ and $j \in {1, 2, \ldots, m}$:
\begin{equation}
    D(i,j) = C_{i,j} + \min_{k \in \{1,\ldots,m\}} \{D(i-1,k) + \pi(A_{i-1,k}, j\})\}
\end{equation}

When computing $D(i,j)$, we consider all possible alignments for the previous target sentence and choose the one that minimizes the sum of the previous cost, the base cost for the current alignment, and the ordering penalty based on the full alignment history.
Note, that the base-case here is $D(0,j) = 0$ for all $j \in \{1, 2, \ldots, m\}$.

\noindent \textbf{Total Cost:} The minimal total hallucination cost is given by \( \min_{j \in \{1,\ldots,m\}} D(n,j) \), and the optimal matching between the source and target sentences is denoted by \( A_{n,j^*} \), where \( j^* \) is the source index that achieves the minimum cumulative hallucination cost.  

However, different VideoLLMs can generate captions of varying lengths for the same video, making direct comparisons of cumulative hallucination cost unfair, as longer captions face higher penalties. To address this, we normalize the cost by computing the maximum possible hallucination cost for a given video, VideoLLM pair. This maximum cost is given by \( (n - d) + \lambda \cdot \frac{d(d-1)}{2} \), where \( d \) represents the number of dynamic-action targets with entailment. The first term accounts for the worst-case base cost when all non-entailed or non-dynamic targets receive the maximum penalty. The second term represents the highest possible ordering penalty, which occurs when all dynamic actions are perfectly inverted. We then define the normalized hallucination metric, i.e., \textbf{\hscoreh}, defined as the ratio of the observed cost to the maximum possible cost, as follows:
\begin{equation}
    \text{\hscoreh: } \Big( \frac{\min_{j \in \{1,\ldots,m\}} D(n,j)}{(n - d) + \lambda \cdot \frac{d(d-1)}{2}} \Big) \cdot 100.
\end{equation}

\paragraph{Omission Cost:}  
Omission refers to information present in the human-annotated caption that is missing from the model-generated caption.  
\textbf{The normalized omission cost, \textbf{\hscoreo}, is assigned just by reversing the roles of the source ($S$) and target ($T$). We then assess whether each line in the human caption can be entailed by the LLM-generated caption. } 
If a human-annotated sentence is not entailed, it indicates that the model-generated caption has omitted that information.
\begin{figure*}[!htp]
    \centering
    \includegraphics[width=\linewidth]{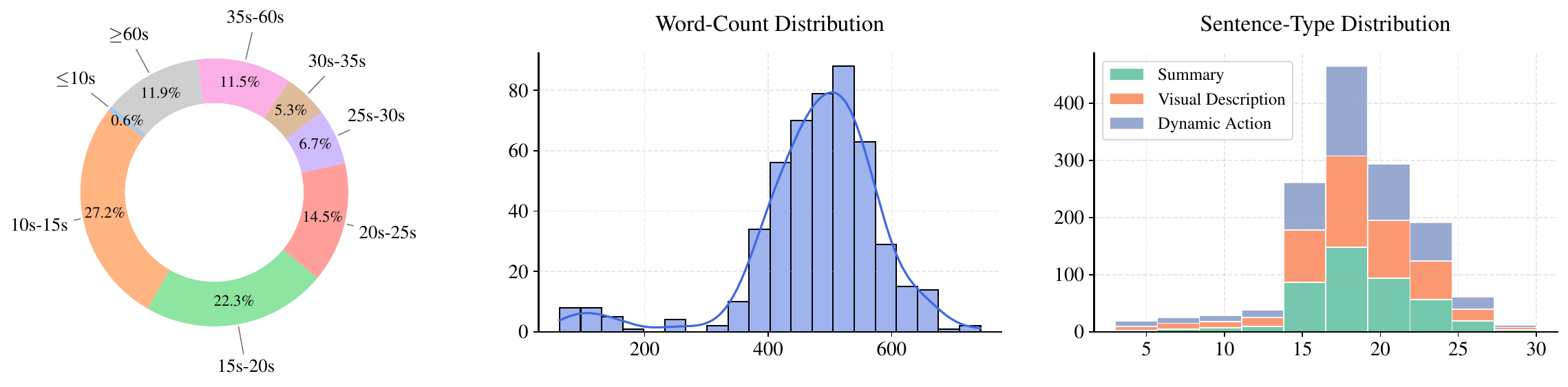}
    \vspace{-2em}
    \caption{\textbf{\hbench Statistics:} video lengths (left), word-counts (middle), and sentence-length (x-axis) distribution by sentence-type (right). Our dataset has a balanced representation across durations and sentence types, and a high word-count density.}
    \label{fig:benchmark_char}
\end{figure*}

\vspace{-0.5em}
\section{Benchmarking Video-LLMs}
\label{sec:benchmark_char}
We detail how we curated our evaluation dataset and describe the evaluations conducted on several open- and closed-source VLMs, along with insights derived from the \hscore trends.

\vspace{-0.5em}
\looseness=-1
\subsection{\hbench: Curation and Statistics}
We collect 500 videos along with their corresponding dense-caption pairs from three sources. First, we utilize existing video understanding datasets \cite{chai2024auroracap} that already contain captions. These videos are manually verified by human authors, and received well in the community.  
Second, we incorporate text-to-video generation datasets \cite{feng2024tc}, which include reference videos and short prompts. Since these prompts are insufficient for dense captioning, we manually annotate $10$ such videos. Lastly, the authors curate additional videos from publicly available sources, such as YouTube, under Creative Commons licenses. We curate $30$ such videos, and also manually annotated, with cross-validation among the authors.

\looseness=-1
In~\cref{fig:benchmark_char} we provide a statistical analysis of our dataset, including our video length distribution, where nearly a quarter of the videos between 15 and 20 seconds, while 12\% exceed 60 seconds. 
\cref{fig:benchmark_char}-middle shows the distribution of word counts (x-axis) in human captions from \hbench, highlighting the rich density of our textual descriptions, averaging 477 words per video and 24.4 words per second. Additionally, \cref{fig:benchmark_char}-right visualizes the distribution of sentence lengths (x-axis) in human captions from \hbench, illustrating how different sentence types contribute to each length category. For instance, we observe approximately 300 sentences with a length of around 20 words, which appear to be fairly evenly distributed across the three sentence categories.
Unlike some existing datasets that over-represent one category, our benchmark maintains a balanced distribution, with approximately equal parts falling into each sentence type. We provide additional details on the types of video clips in our benchmark in~\cref{appendix_sec:addn_benchmark_details}.
Our sentence-level evaluation provides a much finer-grained analysis, averaging $\approx$19 evaluations per video compared to 1-2 questions per video in the baseline benchmarks.
\begin{figure*}
    \centering
    \includegraphics[width=0.9\linewidth]{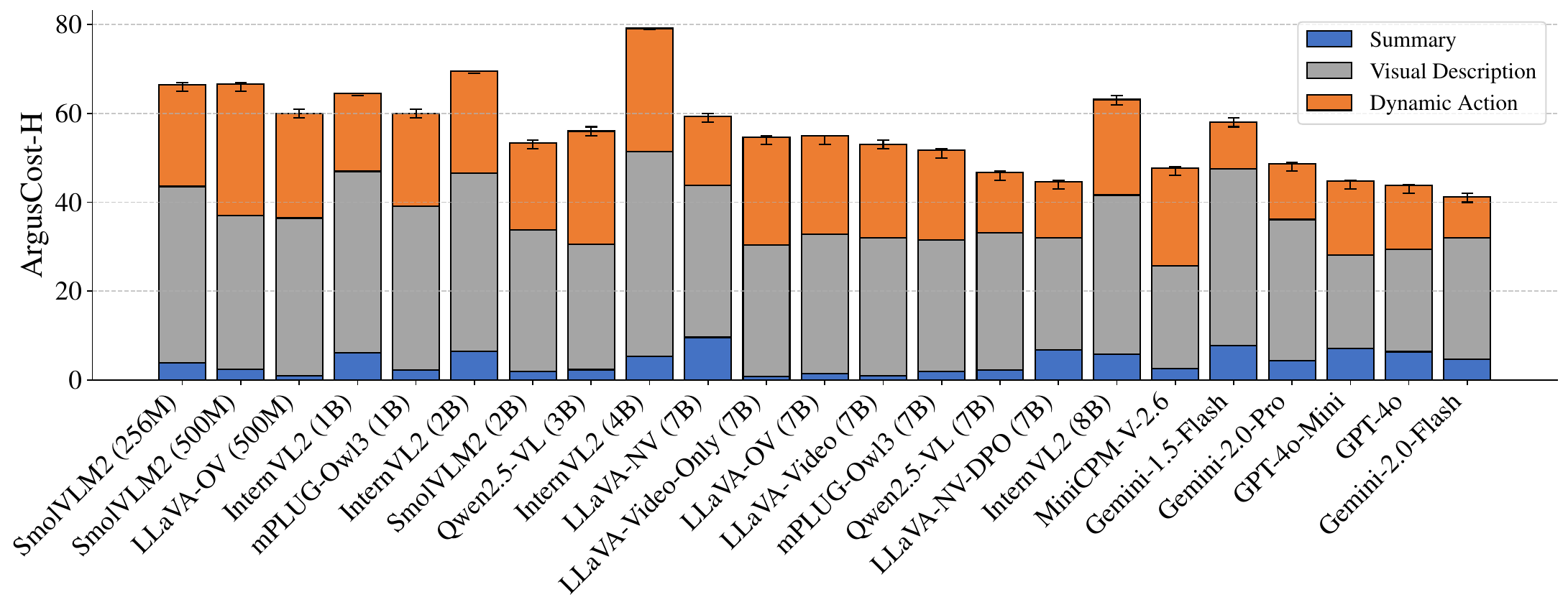}
    \vspace{-1em}
    \caption{\textbf{\hscoreh (Hallucination Cost) across Video-LLMs.} Even top performers like Gemini-2.0-Flash produce up to 40\% hallucinated content. Although summary errors are low, stronger models still fabricate visual details despite improved dynamic action descriptions. Open-source models are ordered by size along the x-axis. Lower \hscoreh is better.}
    \label{fig:hallucination_main_results}
    \vspace{-10pt}
\end{figure*}

\begin{figure*}
    \centering
    \includegraphics[width=0.9\linewidth]{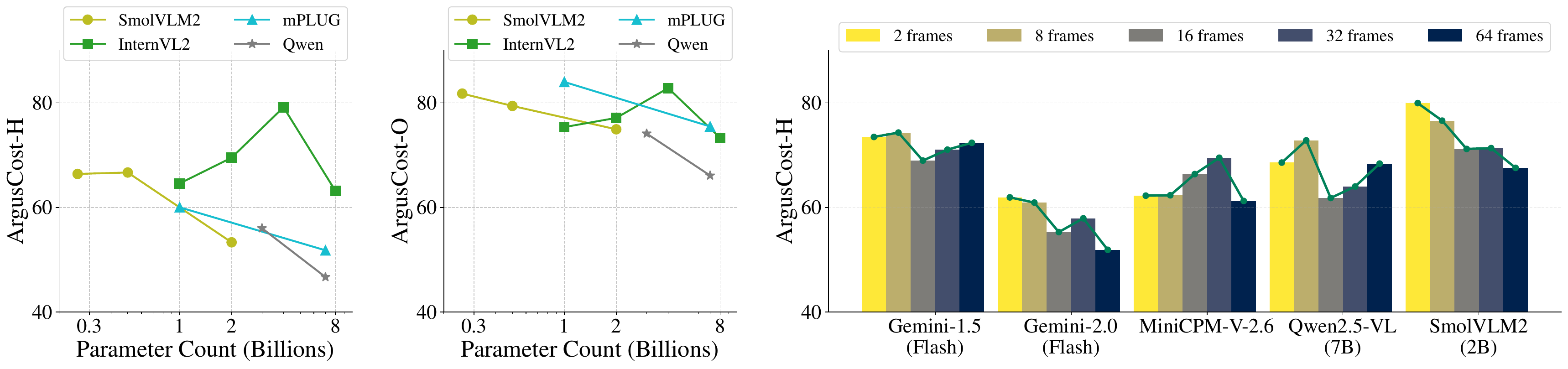}
    \caption{(Left,Middle) \textbf{Relationship between hallucination/omission and model-size.} For most open-source model families (except InternVL2), larger models tend to have lower hallucination (\hscoreh) and omission (\hscoreo) costs, indicating that scale generally improves performance. (Right) \textbf{Relationship between hallucination and frames of the video provided.} Gemini models and SmolVLM2 show a consistent reduction in hallucinations with more frames, while MiniCPM-V and Qwen2.5-VL exhibit fluctuating levels of hallucinations (as measured by \hscoreh). Please see Appendix~\ref{appendix_sec:addn_hall_cost_results} for results on the effect of frames provided on \hscoreo.}
    \label{fig:eff_total_frames}
    \vspace{-1em}
\end{figure*}

\subsection{Evaluations \& Insights}
\label{sec:results}

In this section, we evaluate a range of closed-source and open-source Video-LLMs using the \horus framework, and examine performance trends along with key challenges faced by different models. 
In total, we evaluate $23$ video models, including $18$ open-source models and $5$ state-of-the-art proprietary ones. We sample a total of 16 frames per video (unless specified otherwise) to evaluate the models. If a model has a predefined frame sampling strategy, we adapt it else we apply uniform sampling. Further details on the evaluation setup, including model checkpoints, computational and monetary costs, evaluation prompts, etc, are provided in~\cref{appendix_sec:eval_details}.

\vspace{-10pt}
\paragraph{Video-LLMs have a hallucination problem.}
We present the \hscoreh--the normalized hallucination costs--for different models in~\cref{fig:hallucination_main_results}, where we note that even the best-in-the-market Video-LLMs generate hallucinated content frequently, with significant cost variation across models and training strategies.
While Gemini-2.0-Flash achieves the lowest \hscoreh at 41\%, outperforming both GPT-4o and GPT-4o-mini (each at approximately 44\%), its larger counterpart, Gemini-2.0-Pro, performs worse at 48\%. 
In contrast, the difference in scale between GPT-4o and GPT-4o-mini has little impact on \hscoreh. The Gemini-2.0 series also shows a clear improvement over the 1.5 series, with Gemini-1.5-Flash trailing behind at 58\%.

Among open-source models, LLaVA-Next-Video (DPO) achieves the best result with a \hscoreh of 45\%, closing the gap with proprietary models. However, its non-DPO version performs significantly worse at 59\%, highlighting the potential role of reinforcement learning-based post-training strategies in reducing hallucination. Other strong performers in the open-source category include Qwen 2.5-VL (7B) and MiniCPM-V-2.6, both with \hscoreh below 50\%. On the other hand, the worst-performing model is InternVL2 (4B), with a \hscoreh of 80\%, indicating that smaller models are not inherently more prone to hallucination. In fact, some smaller models, such as SmolVLM2 (2B) and Qwen 2.5-VL (3B), perform substantially better than their peers and even some larger models, with \hscoreh around 55\%.

\cref{fig:hallucination_main_results} also breaks down \hscoreh by sentence type, showing that ``summary" sentences contribute minimally--likely because each caption contains only a few, which models generally handle well. However, we observe that proprietary models like Gemini-2.0-Flash and GPT-4o exhibit slightly higher summary sentence errors than some weaker open-source models. We qualitatively examined these errors and found that these models tend to generate abstract interpretations of motives, atmosphere, and mood, especially when such information is subjective and cannot be reliably grounded. Errors in visual descriptions and dynamic actions are fairly evenly distributed among poorly performing models. However, as we go from models with high to low \hscoreh, errors in dynamic actions decrease more rapidly than in visual descriptions, suggesting that stronger video models are more reliable at describing temporal events than at avoiding fabricated visual details. We present additional results on \hscoreh breakdown by verdict-type (contradiction v/s underdetermined), and cost-type (base-cost v/s order penalty) in~\cref{appendix_sec:addn_hall_cost_results}.

\vspace{-1em}
\paragraph{Do hallucination rates correlate with omission rates?}
\looseness=-1
Two competing factors govern the relationship between hallucinations and omissions. First, overall stronger models may have lower hallucination rates and also reduced \hscoreo. Alternatively, weaker models may tradeoff hallucination and omission by being more (or less) verbose, thus causes an inverse relationship between the two.  
We observed both of these trend types when we analyzed the relationship between hallucination and omission across model families, as shown in~\cref{fig:hallucination_v_omission_model_familiy}. We find a moderately high Pearson correlation of 0.65, supporting the first hypothesis. Most models lie above the $y=x$ line, indicating they omit more than they hallucinate. Notably, despite a significant improvement in \hscoreh for Gemini-2.0-Flash (41\%) compared to Gemini-2.0-Pro (48\%),
its \hscoreo remains around 60\%. In contrast, GPT-4o and GPT-4o-mini have similar \hscoreh (44\%), but GPT-4o exhibits a 5\% lower \hscoreo. LLaVA-Next-Video (DPO), which performed close to proprietary models in hallucination (45\%), has a much higher \hscoreo (85\%), suggesting the model plays it safe and avoids generating uncertain content, often missing important details. In comparison, Qwen 2.5-VL (7B) and MiniCPM-V-2.6 achieve both low \hscoreh and \hscoreo, making them better opensource choices. 
In~\cref{fig:hallucination_v_omission_model_familiy}, marker size indicates average number of sentences in captions generated by each of these models; we do not see any trends corresponding to this — \textit{longer captions do not necessarily lead to higher or lower hallucination}. When we perform a correlation analysis, we find no correlation ($0.09$) between model-generated (average) caption length and \hscoreo, while we observe a low positive correlation ($0.32$) between caption length and \hscoreh.
We also investigate which other characteristics of the video or the human-generated caption may correlate with hallucination and \hscoreo. For instance, we find that both \hscoreh and \hscoreo have mild positive correlation ($\approx 0.25$) with the clip duration (in seconds). We present additional results in the~\cref{appendix_sec:addn_hall_cost_results} due to space constraints.

\vspace{-1em}
\paragraph{Does scale help reduce hallucinations and omissions?}
In our discussion of~\cref{fig:hallucination_main_results}, we showed that smaller Video-LLMs do not necessarily hallucinate more and can sometimes even outperform larger models. However, that analysis grouped all models together, meaning factors like architecture, training data, and other differences would have influenced the results—not just scale. To better isolate the effect of the model scale, we analyze \hscoreh and \hscoreo within 4 model families:  SmolVLM2~\cite{huggingfaceSmolVLM2Bringing}, mPLUG~\cite{ye2024mplug}, Qwen~\cite{yang2024qwen2}, and InternVL2~\cite{chen2024expanding} with sizes ranging from sub-billion to 8 billion parameters
in~\cref{fig:eff_total_frames}~(Left, Middle). We find that for most model families, increasing scale improves performance—both \hscoreh and \hscoreo decrease. For example, in the SmolVLM2 family, scaling from 256 million to 2 billion parameters improves hallucination performance by over 15\% and omission performance by 5\%. Similarly, Qwen 2.5-L improves by 10\% for hallucination and 8\% for omission. However, this trend does not hold for the InternVL2 family, where \hscoreh and \hscoreo initially increase from 1B to 4B parameters before decreasing at 8B, ultimately returning to the level of the 1B-parameter model. It remains an open question what exact components in the InterVL2 training pipeline led to the emergence of such behavior since the smallest InternVL2 (1B) is on par with other smaller 256M and 500M models. We leave this investigation for future work.

\looseness=-1
\vspace{-1.4em}
\paragraph{Effect of frame sampling rate}
In previous experiments, we sampled 16 frames per video for each model since the smallest maximum frame limit among the models in our pool was 16. However, some models can handle up to 64 frames or more, so we now investigate how increasing the total number of frames affects \hscoreh and \hscoreo. 
We perform this ablation on a subset of 50 randomly sampled videos from our benchmark.
We present the results in~\cref{fig:eff_total_frames}. We find that Gemini models and SmolVLM2 consistently improve in both \hscoreh and \hscoreo as the number of frames increases, suggesting that these models effectively leverage additional frame information to generate more reliable captions—reducing both fabricated and omitted information. For example, the \hscoreh for Gemini-2.0-Flash improves by 10.02\% when increasing from 2 to 64 frames, while SmolVLM2 improves by 12.38\%. However, for some open-source models, such as MiniCPM-V-2.6 and Qwen2.5-VL, the effect on \hscoreh is less clear. MiniCPM-V-2.6 initially worsens as the number of frames increases but then drops back to a similar level as with 2 frames at 64 frames. Qwen2.5-VL follows an inverse pattern, where hallucination first improves but then worsens at 64 frames. For omission, however, all models show consistent improvement as the number of frames increases. 
We present the detailed omission results with performance on all frames in~\cref{appendix_sec:addn_hall_cost_results},  due to space constraints.
This suggests that while using more frames can help models capture more information and reduce omissions, models may also struggle to balance relevant and irrelevant details, leading to greater reliance on learned patterns, abstraction, or speculative inferences rather than staying grounded in the actual video content. Additionally, attention and memory limitations can force models to summarize or interpolate details inaccurately, further increasing the risk of hallucinations.

\begin{table}[t]
\centering
\small
\begin{tabular}{@{}l@{\hspace{0.5em}}c@{\hspace{0.5em}}c@{}}
\toprule
\textbf{Model} & \multicolumn{2}{c}{\textbf{\hscoreh}} \\
\cmidrule(l){2-3}
 & \textbf{Intra-Prompt} & \textbf{Inter-Prompt} \\
\midrule
SmolVLM2 & 72.7 $\pm$ 2.9 & 73.0 $\pm$ 3.0 \\
InternVL2 & 83.4 $\pm$ 1.6 & 83.4 $\pm$ 1.6 \\
MiniCPM-V-2.6 & 69.2 $\pm$ 2.4 & 67.0 $\pm$ 2.4 \\
Qwen2.5-VL & 62.1 $\pm$ 2.8 & 62.9 $\pm$ 2.9 \\
\bottomrule
\end{tabular}
\caption{\textbf{Sensitivity to the Prompt.} We sample multiple dense captions by varying prompt parameters and compute \hscoreh. The low standard deviation of \hscoreh across many Video-LLMs demonstrates its robustness.}
\label{tab:combined_prompt}
\vspace{-10pt}
\end{table}

\begin{figure}
    \centering
    \includegraphics[width=0.85\linewidth]{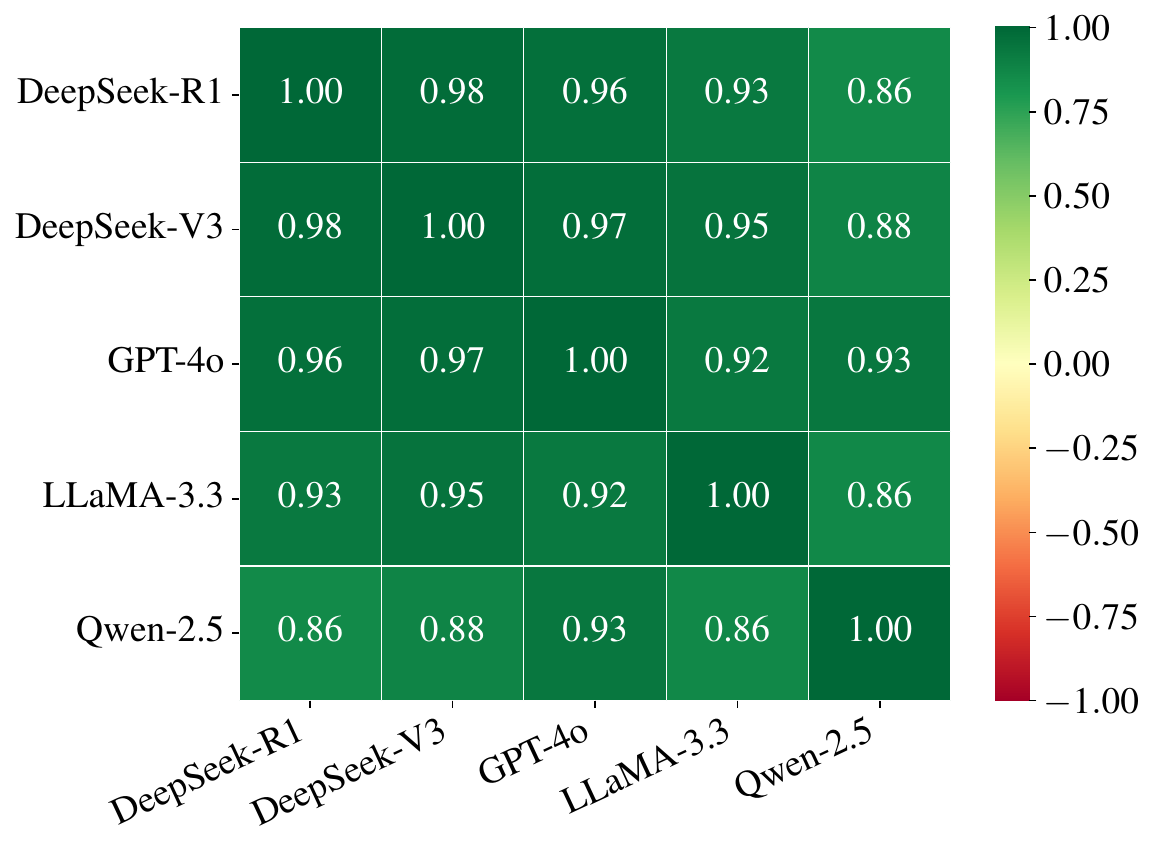}
    \caption{\textbf{Sensitivity to different LLMs-as-judge.} High Pearson ranking correlations across different LLMs-as-judge indicate strong agreement in evaluations and robustness in ranking trends.}
    \label{fig:judge_sensitivity}
    \vspace{-8pt}
\end{figure}

\subsection{\horus Sensitivity Analysis}
\label{subsec:sensitivity_analysis}
Since one of the important steps of our evaluation process involves inherent stochasticity—such as the choice of prompt for generating captions—one concern is whether the observed trends generalize. To address this, we conduct a sensitivity analysis on a subset of $50$ randomly sampled clips, evaluating intra-prompt variation, inter-prompt variation, and the choice of LLM-as-judge. 

\looseness=-1
For intra-prompt analyses, we sample three captions for each model using our default prompt (``Describe the video in great detail.") at the default decoding temperature and analyze the variation in \hscoreh. For inter-prompt analyses, we run experiments with three additional prompts: ``Explain the content of this clip thoroughly.", ``Can you summarize all the key elements and events of this video?", and ``Walk me through this video scene by scene.", and analyze the variation in \hscoreh. ~\cref{tab:combined_prompt} shows low standard errors in the range of $1$–$3$ for all model settings. As expected, the error is slightly higher in inter-prompt setting compared to the intra-prompt setting.

\looseness=-1
Another potential source of variation is the choice of LLM used as the judge for generating NLI judgments, as the entire evaluation process depends on it. By default, we use GPT-4o. However, since GPT-4o is also one of the models being evaluated, there is a possibility of self-bias. To investigate this, we conducted experiments using four additional LLM-as-judge models from different families and sizes, all of which are strong in NLI evaluation: DeepSeek-R1~\cite{guo2025deepseek}, DeepSeek-V3~\cite{liu2024deepseek}, LLaMa-3.3~\cite{grattafiori2024llama}, and Qwen-2.5~\cite{yang2024qwen2}. ~\cref{fig:judge_sensitivity} presents the Pearson ranking correlations $r$ between rankings produced by different judge models. We find that these correlations are very high ($r \geq 0.92$), indicating strong agreement across judge models. Notably, GPT-4o (our default judge) has a ranking correlation of $r = 0.96$ with DeepSeek-R1, $r = 0.97$ with DeepSeek-V3, $r = 0.93$ with LLaMa-3.3, and $r = 0.92$ with Qwen-2.5, suggesting that our evaluation remains robust when a frontier judge model is used.
A detailed discussion and additional sensitivity analyses are provided in~\cref{appendix_sec:addn_sensitivity_analysis}.
Additionally, we conducted a human study to assess the reliability of our LLM-based evaluation. Participants viewed videos and judged whether they agreed with the LLM’s line-by-line verdicts on model-generated captions. We observed a high average agreement rate of $91.26\%$, with most disagreements arising from fine-grained visual details. Further methodology and analysis are provided in Appendix~\ref{appendix_sec:human_study_details}.
\section{Discussion \& Conclusion}
\label{sec:conclusion}

In this work, we present \horus, a first-of-its-kind evaluation framework for quantifying hallucinations and omissions in dense video captioning, addressing key limitations of previous QA-based approaches. We propose dual metrics, \hscoreh and \hscoreo, to assess the accuracy and completeness of generated captions. Our experiments show that even top VideoLLMs, such as Gemini-2.0-Flash, produce a significant amount of hallucinated content, highlighting the gap between targeted verification and open-ended generation. Moreover, these trends are consistent across multiple samplings, varied input prompts, and different LLMs-as-judges. These findings emphasize the need for future research to mitigate hallucinations and improve the overall accuracy of dense video captions.

\section*{Acknowledgments}
The authors would like to thank Yuxin Wen, Sachin Shah, Sean McLeish, Udit Chugh, Prachi Rawal, and many others who helped us with the human study.

This work was made possible by DARPA TIAMAT and the NSF TRAILS Institute (2229885). Commercial support was provided by Capital One Bank, the Amazon Research Award program, and Open Philanthropy.

{
    \small
    \bibliographystyle{ieeenat_fullname}
    \bibliography{main}
}

\clearpage
\appendix
\onecolumn
\centering

{\large \textbf{\horus: Hallucination and Omission Evaluation Framework for Video-LLMs}} \\
\vspace{0.5em}{\large Appendix} \\
\vspace{1.5em}
\raggedright

\section{The Limitations of Current QA Approaches - Extended}\label{appendix_sec:method_sub:current-approaches}
Current evaluation strategies predominantly rely on a question-answering (QA) paradigm, where the model is tested with two types of questions: one basic type that expects a ``Yes" (confirming the presence of information in the video) and another  ``leading/hallucinated" type that expects a ``No" (flagging potential hallucinations). Although this setup simplifies evaluation, it suffers from several critical limitations.

\paragraph{Lack of Dependency on Over Visual Understanding}
A key limitation of binary QA-based evaluation is that models can sometimes answer questions correctly using general world knowledge alone, without processing the visual input \cite{lin2023revisiting, zohar2024apollo}. To demonstrate that this issue persists in current video hallucination datasets, we conduct an experiment where an LLM is tested on QA metrics without access to visual context. High performance in this setting indicates that these evaluations do not reliably measure visual grounding. Specifically, we use GPT-4o to answer a randomly selected subset of questions from Video-HalluCer \cite{wang2024videohallucer} using only the textual question. Surprisingly, the model correctly answers 32.52\% of basic hallucination-related question pairs, with some subcategories (\texttt{external\_nonfactual\_instruct}) reaching 61.33\%, despite a chance performance of just 25\%.

\paragraph{Verification Ability Does Not Equate to Strong Generation}
Video LLMs are used for a range of tasks, including question-answering (QA) and open-ended generation, such as captioning and summarization. While QA evaluates a model’s ability to verify information, it does not necessarily reflect its capability to generate accurate and coherent descriptions. Recent work on hallucination evaluation in image captioning \cite{kaul2024throne} suggests that proficiency in fact verification does not always correlate with strong open-ended generation and may even be inversely related. We observe a similar pattern in the qualitative example shown in Figure~\ref{fig:teaser_figure}, where a Video-LLM correctly verifies that there is only one chameleon in a video clip but mistakenly identifies it as two distinct chameleons in free-form generation. This highlights the need for a dedicated benchmark to assess hallucinations specifically in open-ended generation. 
To further quantify this behavior, we conducted an experiment using 50 randomly selected videos from our dataset. For each video, we extracted each line (one at a time) from a model’s generated caption and reformulated it into a verification prompt, which was then presented to the same model along with the visual video input:

\begin{quote}
\textbf{Evaluate the accuracy of this statement about the video content:} \\
\textbf{Statement:} \texttt{\{model-caption-line\}} \\
(a) True \\
(b) False \\
Respond with only the letter of your evaluation: (a) or (b).
\end{quote}

\begin{figure}
    \centering
    \includegraphics[width=0.65\linewidth]{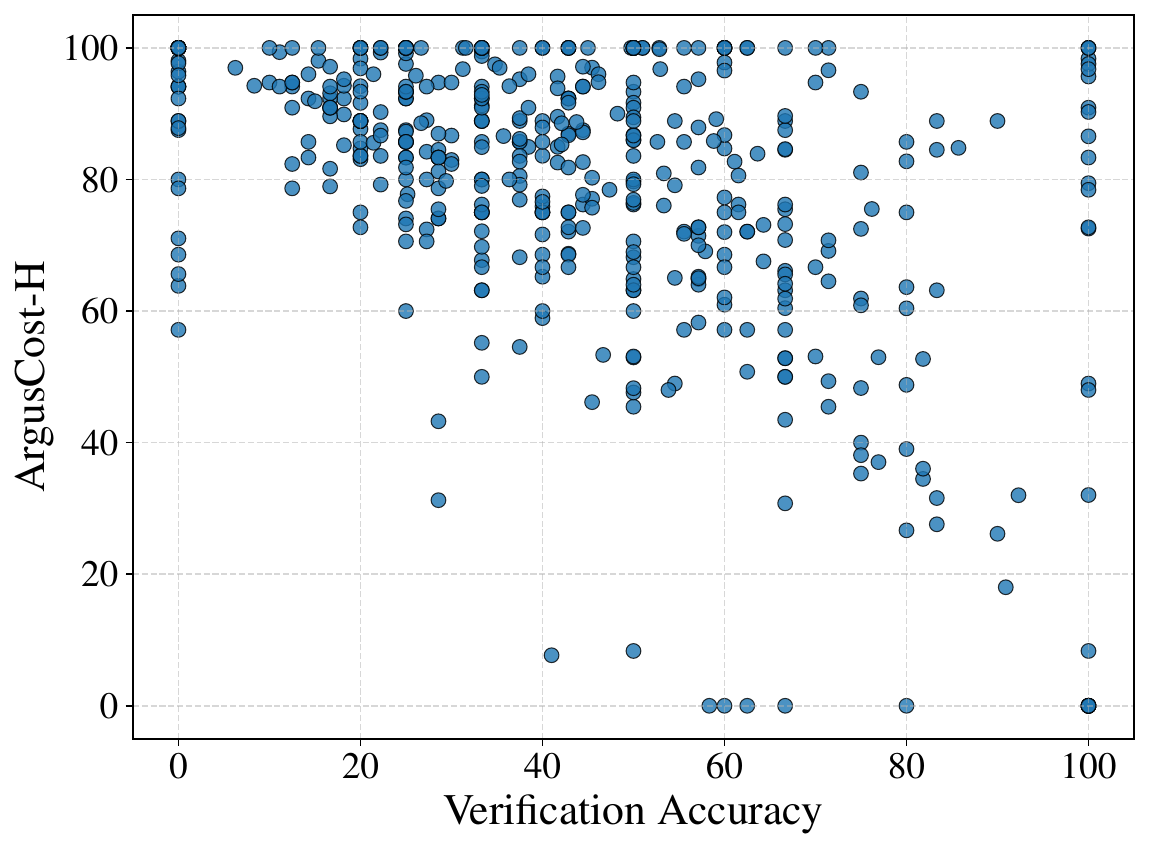}
    \caption{Relationship between verification accuracy and hallucination cost across 50 video samples.}
    \label{fig:correlation_ver_gen}
\end{figure}

Using this setup, we measured the \textit{verification accuracy} of the model. Since we already had ground-truth entailment labels from our captioning evaluation--indicating whether a given line was factual or hallucinated--we could compare them against the model’s verification responses. This allowed us to analyze the relationship between a model’s verification accuracy and its hallucination rate. Because both tasks were performed by the same model, a strong negative correlation between verification accuracy and hallucination cost would suggest that these abilities are intertwined. In contrast, a weak correlation would imply that a model may accurately verify information while still hallucinating in free-form generation, or vice versa. In our analysis, we observed a Pearson correlation of $-0.48$ (see~\cref{fig:correlation_ver_gen}), indicating a moderate negative correlation. These findings further highlight the need for a dedicated benchmark to assess hallucinations specifically in open-ended generation tasks.

\paragraph{Restricted Error Coverage Due to Predefined Scope}
Another limitation of the QA-based approach is its restricted scope; the predefined questions cover only a narrow range of possible errors. As a result, the full spectrum of hallucinations remains hidden until the model is allowed to generate free-form captions.

\paragraph{Inability to Capture Multi-Event Hallucinations}
Many videos consist of multiple interrelated events, yet QA-based evaluation methods typically focus on isolated, short events, if at all. This narrow focus fails to account for the complex interactions and temporal dependencies between events, which can contribute to more severe hallucinations. For example, if a video contains four sequential events (event-a, event-b, event-c, and event-d), a binary QA approach would require separately probing the existence and order of each event. Moreover, with a 50\% chance of guessing correctly in a binary setting, it raises questions about whether the model genuinely understands the content. In contrast, our free-form generation strategy allows the model to produce captions based on its interpretation of the video, enabling a more direct evaluation of its accuracy.

\section{Extended Related Work}
\label{appendix_sec:extended_related_work}

\textbf{Natural Language Inference}
Natural Language Inference (NLI) is a classic natural language processing task aimed at determining whether a given hypothesis can logically be inferred from a provided premise \cite{bowman2015snli,camburu2018snli,chen2021neurallog,williams2020anlizing,demszky2018transforming,williams2017broad}. Formally, NLI involves classifying the relationship between a hypothesis and a premise into one of three categories: entailment (hypothesis logically follows from the premise), contradiction (hypothesis contradiction implies the hypothesis), or neutrality (no direct logical relation between the two). 
As one of the earliest works, Bowman et al.~\cite{bowman2015snli} released the SNLI corpus, a large-scale annotated datasets for NLI. SNLI consists of over 500k sentence pairs, each labeled to indicate entailment, contradiction, or semantic independence, enabling training of modern machine learning models. Williams et al.~\cite{williams2017broad} introduced MNLI, a large scale NLI dataset extending the traditional natural language inference task to multiple-genres of spoken and written english such as reports, letters, fiction works, etc. Demszky et al.~\cite{demszky2018transforming} proposed an automated way of generating NLI datasets based on existing question-answering datasets.
More recently, NLI techniques have been applied to evaluating factuality in domains such as natural language summarization by analyzing the relationship between a generated summary and the source document \cite{scire2024fenice,gekhman2023trueteacher}.

In this work, we adapt the broader NLI-framework to assess hallucination and omission in free-form generation by Video-LLMs. Specifically, we frame this evaluation as a textual NLI problem, treating the ground truth human caption as the premise (and hypothesis in omission evaluation) and the generated model caption as the hypothesis (and the premise in omission evaluation).

\noindent\textbf{Dense Video Captioning.} Video Captioning is one of the traditional tasks since the dawn of deep learning. Dense video captioning is a fine-detail oriented subtask in which the model has to annotate all the important visual and temporal details in the video. Some of the common applications include surveillance monitoring, video content retrieval, and educational accessibility. Additionally, dense video captioning can generate synthetic data for training video-generative models since dense captioning by humans at scale can be quite expensive. Most of the early works~\cite{xu2019joint,mun2019streamlined,wang2021end,kim2024you,iashin2020better,iashin2020multi,deng2021sketch,zhou2024streaming, li2018jointly,qasim2025dense} handled this via combination of multiple video encoders and LSTM-based or transformer language decoders and have shown good results on highly specific academic datasets. These pioneering approaches laid the groundwork for the task by demonstrating that integrating various temporal and spatial features was key to producing accurate and meaningful captions. However, the reliance on narrowly focused datasets often limited the generalizability of these systems in more diverse, real-world scenarios. Recent breakthroughs have led to the emergence of generalist dense video captioning capabilities, driven by large-scale multimodal models such as Gemini~\cite{team2023gemini} and GPT-4v~\cite{achiam2023gpt} and others. These modern approaches rely on end-to-end training or fine-tuning expansive transformer-based decoder models, augmented with specialized video encoders like SigLIP~\cite{zhai2023sigmoid}, CLIP~\cite{radford2021learning}, Video-MAE~\cite{tong2022videomae}, and Eva-CLIP~\cite{fang2023eva} and on top of internet-scale multi-modal interleaved datasets.
\section{Additional Evaluation Details}
\label{appendix_sec:eval_details}
We use two NVIDIA A40 GPUs, each with 48GB of memory, and two NVIDIA A100 GPUs, each with 82GB of memory, for experiments with open-source models. All models and their checkpoints are listed in Table~\ref{tab:model_checkpoints}. The open-source models are available via \href{https://huggingface.co/models}{HuggingFace}, while the proprietary models are accessible through their respective providers' APIs. All evaluations for the proprietary models were conducted in February 2025.

\begin{table*}[b]
\centering
\begin{tabular}{@{}c|c@{}}
\toprule
\textbf{Model Name}   & \textbf{Checkpoint}                        \\ \midrule
Gemini-1.5-Flash      & \texttt{gemini-1.5-flash}                  \\
Gemini-2.0-Pro        & \texttt{gemini-2.0-pro-exp-02-05}                   \\
Gemini-2.0-Flash      & \texttt{gemini-2.0-flash}                           \\
GPT-4o                & \texttt{gpt-4o}                                     \\
GPT-4o-Mini           & \texttt{gpt-4o-mini}                                \\
InternVL2 (1B)        & \texttt{OpenGVLab/InternVL2-1B}                     \\
InternVL2 (2B)        & \texttt{OpenGVLab/InternVL2-2B}                     \\
InternVL2 (4B)        & \texttt{OpenGVLab/InternVL2-4B}                     \\
InternVL2 (8B)        & \texttt{OpenGVLab/InternVL2-8B}                     \\
LLaVA-OV (7B)         & \texttt{lmms-lab/llava-onevision-qwen2-7b-ov}       \\
LLaVA-OV (500M)       & \texttt{lmms-lab/llava-onevision-qwen2-0.5b-ov}     \\
LLaVA-Video (7B)      & \texttt{lmms-lab/LLaVA-Video-7B-Qwen2}              \\
LLaVA-Video-Only (7B) & \texttt{lmms-lab/LLaVA-Video-7B-Qwen2-Video-Only}   \\
LLaVA-NV-DPO (7B)     & \texttt{lmms-lab/LLaVA-NeXT-Video-7B-DPO}           \\
LLaVA-NV (7B)         & \texttt{lmms-lab/LLaVA-NeXT-Video-7B}               \\
MiniCPM-V-2.6         & \texttt{openbmb/MiniCPM-V-2\_6}                     \\
mPLUG-Owl3 (1B)       & \texttt{mPLUG/mPLUG-Owl3-1B-241014}                 \\
mPLUG-Owl3 (7B)       & \texttt{mPLUG/mPLUG-Owl3-7B-241101}                 \\
Qwen2.5-VL (7B)       & \texttt{Qwen/Qwen2.5-VL-7B-Instruct}                \\
Qwen2.5-VL (3B)       & \texttt{Qwen/Qwen2.5-VL-3B-Instruct}                \\
SmolVLM2 (256M)       & \texttt{HuggingFaceTB/SmolVLM2-256M-Video-Instruct} \\
SmolVLM2 (500M)       & \texttt{HuggingFaceTB/SmolVLM2-500M-Video-Instruct} \\
SmolVLM2 (2B)         & \texttt{HuggingFaceTB/SmolVLM2-2.2B-Instruct}       \\ \bottomrule
\end{tabular}
\caption{Details on model names and corresponding checkpoints. All open-source models are available via \href{https://huggingface.co/models}{HuggingFace}, and proprietary models are available via respective providers.}
\label{tab:model_checkpoints}
\end{table*}
\section{Additional Sensitivity Analysis}
\label{appendix_sec:addn_sensitivity_analysis}

\begin{figure}
    \centering
    \includegraphics[width=\linewidth]{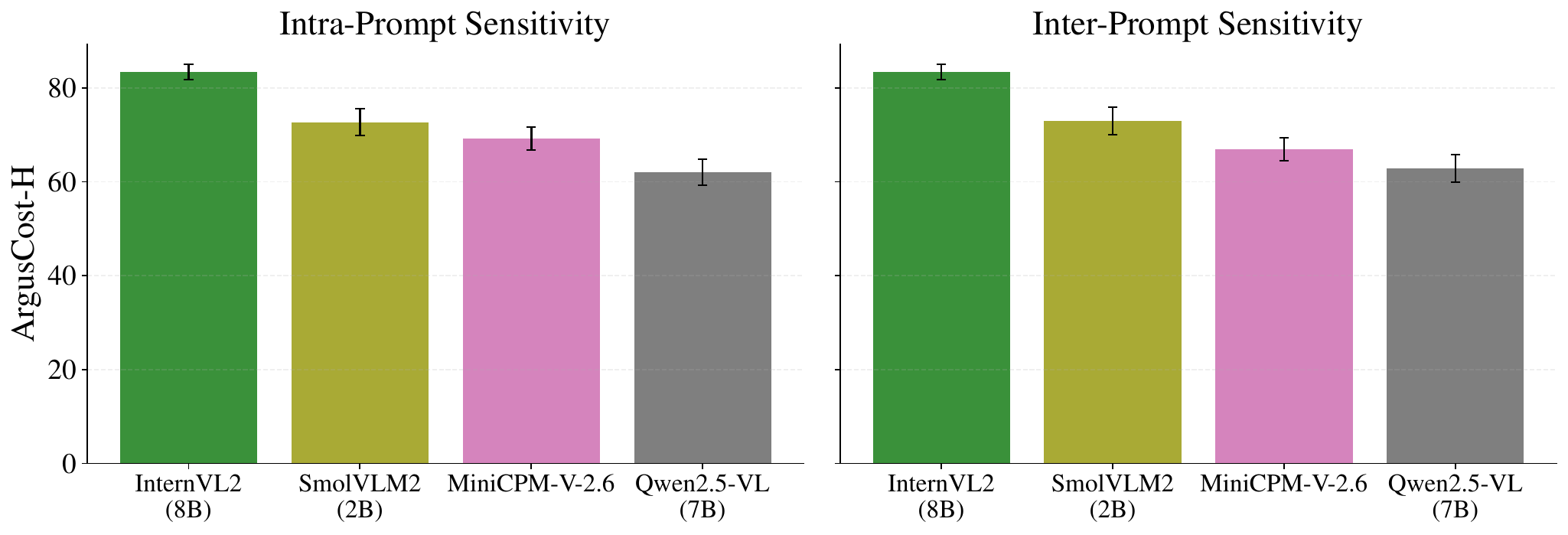}
    \caption{\textbf{Sensitivity to intra- and inter-prompt variations.} Low standard error both intra-prompt (same prompt, default temperature) and inter-prompt (across prompts) suggests consistent hallucination costs across models.}
    \label{fig:intra_inter_prompt_sensitivity}
\end{figure}

\paragraph{Sensitivity to Intra-Prompt Variation}
By intra-prompt variation, we refer to the inherent stochasticity in Video-LLM responses due to decoding at a default temperature. To assess whether this variability significantly impacts our results, we sample multiple captions for each model using our default prompt (“Describe the video in great detail.”) and analyze the variation in hallucination costs. In Figure~\ref{fig:intra_inter_prompt_sensitivity}, we plot the average hallucination cost across three runs along with the standard error. We find that the standard error is minimal, indicating little variation in results. For instance, for InternVL2, we observe a standard error of $1.63$, and for Qwen, the standard error is $2.86$. This suggests that despite stochasticity in decoding, intra-prompt variation does not substantially affect our conclusions.

\paragraph{Sensitivity to Inter-Prompt Variation}
Another source of variation is the specific prompt used to generate captions. By default, we prompt all models with ``Describe the video in great detail." while excluding any special tags required by certain models (e.g., \texttt{<|usertext|>}, etc.). To assess the impact of prompt variation, we run experiments using three additional prompts: ``Explain the content of this clip thoroughly.", ``Can you summarize all the key elements and events of this video?", and ``Walk me through this video scene by scene." Figure~\ref{fig:intra_inter_prompt_sensitivity} reports the hallucination cost averaged over these runs, along with the standard error. As with intra-prompt variation, we find minimal variation across different prompts, indicating that the trends remain stable.

\paragraph{Sensitivity to the LLM-as-judge}
Another potential source of variation is the choice of LLM used as the judge for generating NLI judgments, as the entire evaluation process depends on it. By default, we use GPT-4o. However, since GPT-4o is also one of the models being evaluated, there is a possibility of self-bias. To investigate this, we conducted experiments using four additional LLM-as-judge models from different families and sizes, all of which are strong in NLI evaluation: DeepSeek-R1~\cite{guo2025deepseek}, DeepSeek-V3~\cite{liu2024deepseek}, LLaMa-3.3~\cite{grattafiori2024llama}, and Qwen-2.5~\cite{yang2024qwen2}. Figure~\ref{fig:judge_sensitivity} presents the Pearson ranking correlations $r$ between rankings produced by different judge models. We find that these correlations are very high ($r \geq 0.92$), indicating strong agreement across judge models. Notably, GPT-4o (our default judge) has a ranking correlation of $r = 0.96$ with DeepSeek-R1, $r = 0.97$ with DeepSeek-V3, $r = 0.93$ with LLaMa-3.3, and $r = 0.92$ with Qwen-2.5, suggesting that our evaluation remains robust when a frontier judge model is used.

\begin{figure}
    \centering
    \includegraphics[width=\linewidth]{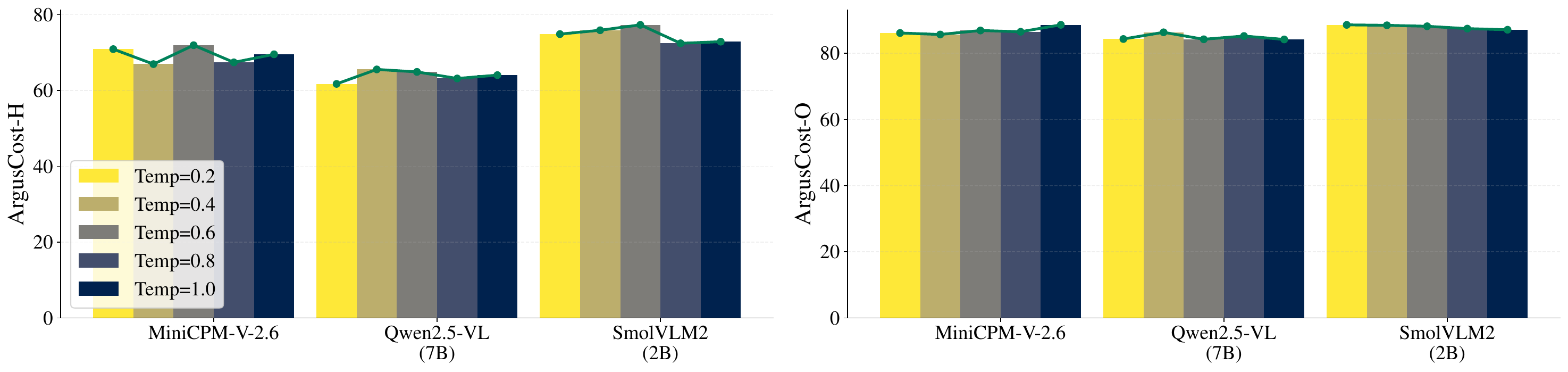}
    \caption{\textbf{Sensitivity across decoding temperatures.} Hallucination costs show little variation across temperatures, indicating low sensitivity to decoding changes}
    \label{fig:temp_sensitivity}
\end{figure}

\paragraph{Sensitivity to Decoding-Temperature}
In Figure~\ref{fig:temp_sensitivity}, we visualize the effect of increasing decoding temperature on hallucination and omission costs. We do not see a clear trend across models here, as the hallucination costs oscillates for MiniCPM-V-2.6 and (to a slightly smaller degree) for Qwen2.5-VL. Whereas, for SmolVLM2, we see a consistent decrease in hallucination and omission cost as the temperature increases.

\section{Human Study Details}
\label{appendix_sec:human_study_details}
While we collect ground-truth human captions, several components of our evaluation pipeline are automated. For example, the NLI-based evaluation uses a large language model (LLM) that compares human and model-generated captions. To ensure reliability, we follow best practices for LLM-as-judge approaches and provide in-context examples. However, differences in detail can still lead to occasional mismatches. A model might include a minor element omitted by the human—either deemed unimportant or overlooked. If that detail isn’t commonly inferred knowledge, the LLM may flag it as a hallucination. For instance, “he is running” versus “he is running fast” is generally acceptable by the evaluator-LLM, as speed is implied. In contrast, mentioning a background flower not noted by the human may be marked as a hallucination. While human captions emphasize salient temporal events, occasional false positives in hallucination detection may still occur.

\begin{figure}
    \centering
    \includegraphics[width=\linewidth]{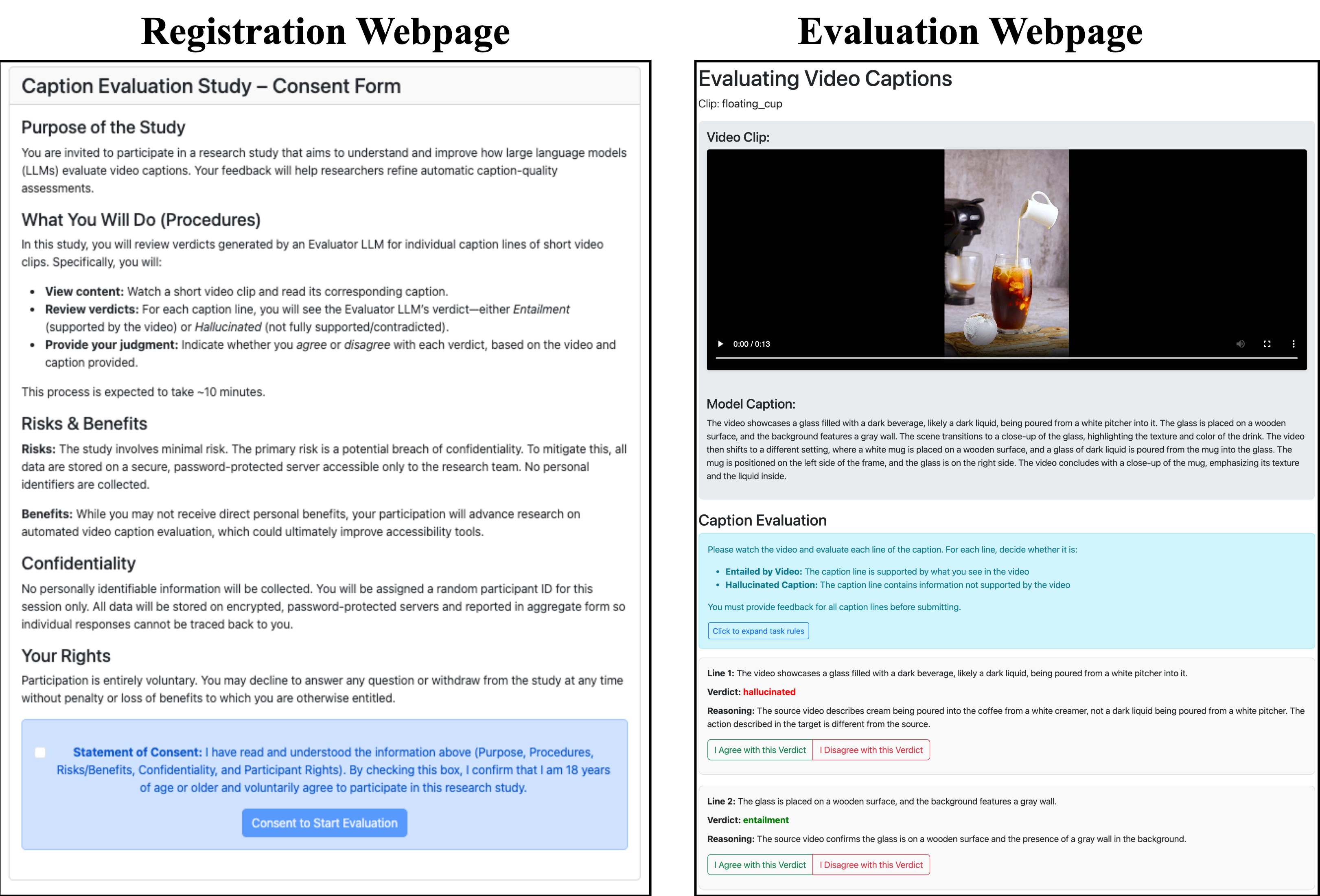}
    \caption{Registration Webpage (Left) and the task webpage (right) from an example video in the human study survey.}
    \label{fig:human_study}
\end{figure}

To assess the reliability of the LLM-based evaluation alongside human annotations, we conducted a human study. Participants were shown the original video along with the NLI evaluator LLM's line-by-line outputs, including whether each segment of the model caption was labeled a ``hallucination." They were then asked to indicate whether they agreed with each verdict. Agreement suggests alignment between the LLM's judgment and human perception, reinforcing the evaluation’s reliability. Disagreement may indicate either limitations in the human caption (e.g., under-specification) or errors in the LLM's reasoning, potentially affecting the final evaluation score. Please see Fig.~\ref{fig:human_study} for details of the registration webpage and the task webpage from an example video in the survey.

\begin{figure}
    \centering
    \includegraphics[width=0.7\linewidth]{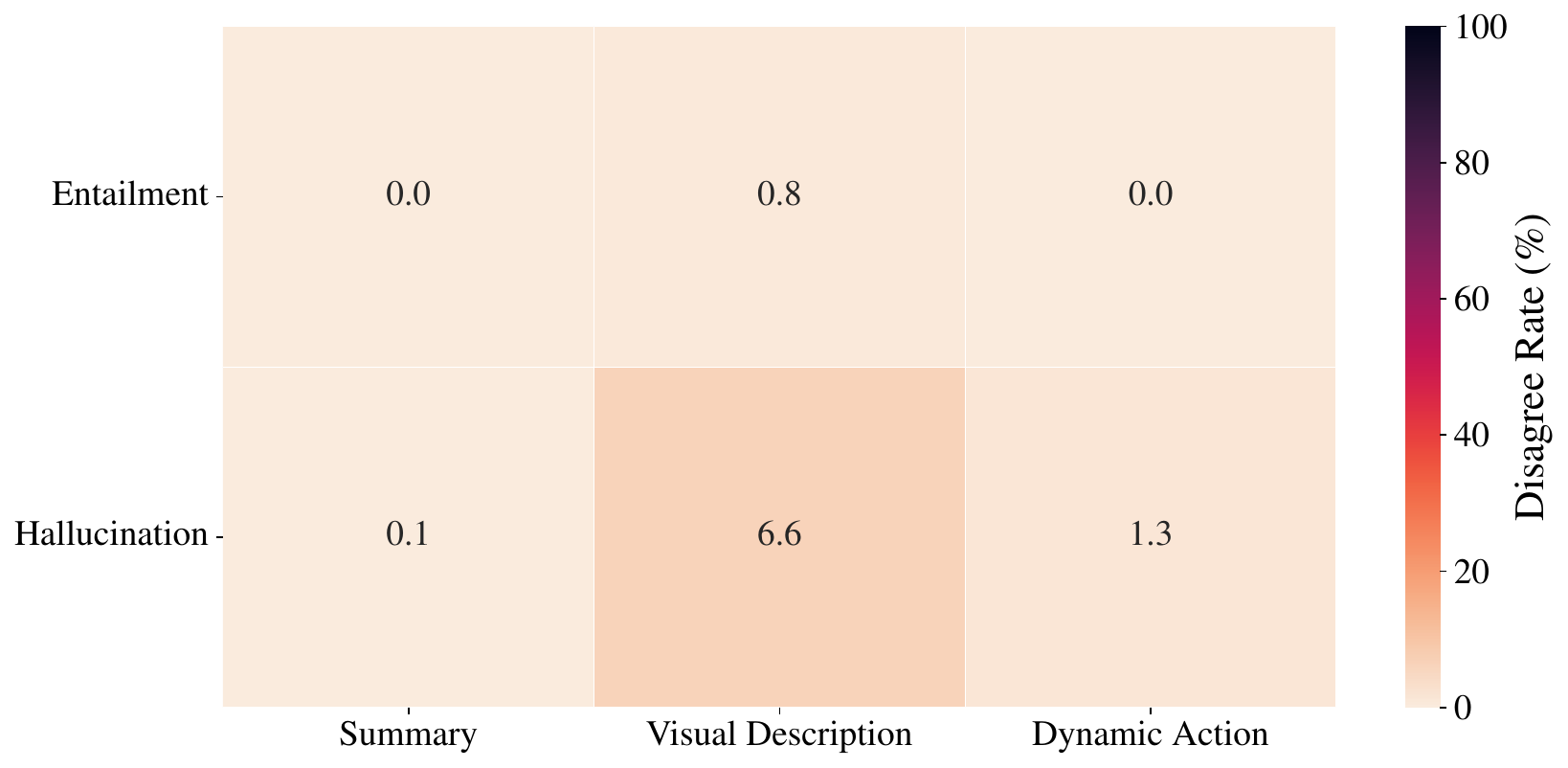}
    \caption{Human-study disagreement rate decomposed into ``LLM-verdict-type" and the ``sentence-type". Overall, disagreement stands at $\sim 9\%$, with the majority contribution being ``Visual-Description" sentences that were classified as ``Hallucination" by the NLI-evaluation LLM.}
    \label{fig:human_study_disagree}
\end{figure}

This study was conducted with $26$ graduate student volunteers. The human study protocol was reviewed and granted exemption by our institution’s Institutional Review Board (IRB). All participants provided informed consent prior to viewing the videos and completing the survey. No personally identifiable information was collected. Each human participant reviewed a minimum of three video clips, with the option to evaluate more; on average, participants assessed $3.54$ clips. We measured the average agreement rate \textit{per clip}—defined as the proportion of lines within a clip for which human reviewers agreed with the LLM evaluator's verdict—and found it to be $91.26\%$. A detailed breakdown of disagreement sources is presented in Fig.~\ref{fig:human_study_disagree}. The highest rate of disagreement occurred for sentences categorized as ``Visual Description" that were labeled as ``Hallucination" by the LLM evaluator but judged as ``Entailment" by human reviewers. 
This pattern is expected, as video-based LLMs could mention fine-grained visual details from both the foreground and background, many of which could be difficult to capture exhaustively in the initial human annotation. Nevertheless, we believe that this is not a major concern, as the overall disagreement rate remains low, and even the strongest models exhibit substantially higher hallucination rates--sometimes reaching up to $40\%$ disagreement. 
Moreover, the fact that most disagreements are skewed toward false positives for hallucination is, in fact, advantageous. It indicates that the evaluation is conservative: lines labeled as ``Entailment" are highly likely to be genuinely grounded in the video content. While this may lead to slight over-penalization for accurate but omitted visual details, such conservatism is well-suited to high-stakes evaluation settings, where ensuring the reliability of entailment judgments is crucial.

\section{Additional Benchmark Details}
\label{appendix_sec:addn_benchmark_details}

\paragraph{Additional Benchmark Curation Details.} Our dataset's diversity also stems from its varied video sources. As discussed earlier, we construct the dataset through manual curation and annotation, supplemented by existing video-understanding and text-to-video generation benchmarks such as VDC \cite{chai2024auroracap} and TC-Bench \cite{feng2024tc}, which themselves integrate diverse video collections. Specifically, 12\% of our dataset comes from Ego4D~\cite{grauman2022ego4d}, which focuses on egocentric perspective videos, while 25\% is sourced from Panda-70M~\cite{chen2024panda}, a dataset of high-resolution YouTube clips spanning various domains, including TV shows, cooking, gaming, and sports. The remaining 63\% is drawn from free stock platforms like Mixkit~\footnote{https://mixkit.co/}, Pexels~\footnote{https://www.pexels.com/}, and Pixabay~\footnote{https://pixabay.com/}, providing high-resolution footage of scenic views and human activities. Additionally, about 4\% is curated from miscellaneous sources, further augmenting the benchmark with a variety of temporal events. We note that for VDC~\cite{chai2024auroracap} the first round of captioning is performed by GPT-4o, which is later manually verified and corrected by the human authors. Since, we use GPT-4o as a judge for our entailment task, there is a possibility of GPT-4o's scores inflating because of self-bias. To ensure, that is not the case, we performed an ablation with many different state-of-the-art LLMs-as-judges in Section~\ref{subsec:sensitivity_analysis}, and found the Pearson ranking correlation to be extremely high.

\section{Additional Evaluation Ablations}
\label{appendix_sec:addn_hall_cost_results}

\paragraph{Cost-Breakdown by Hallucination-Type}
\begin{figure}
    \centering
    \includegraphics[width=\linewidth]{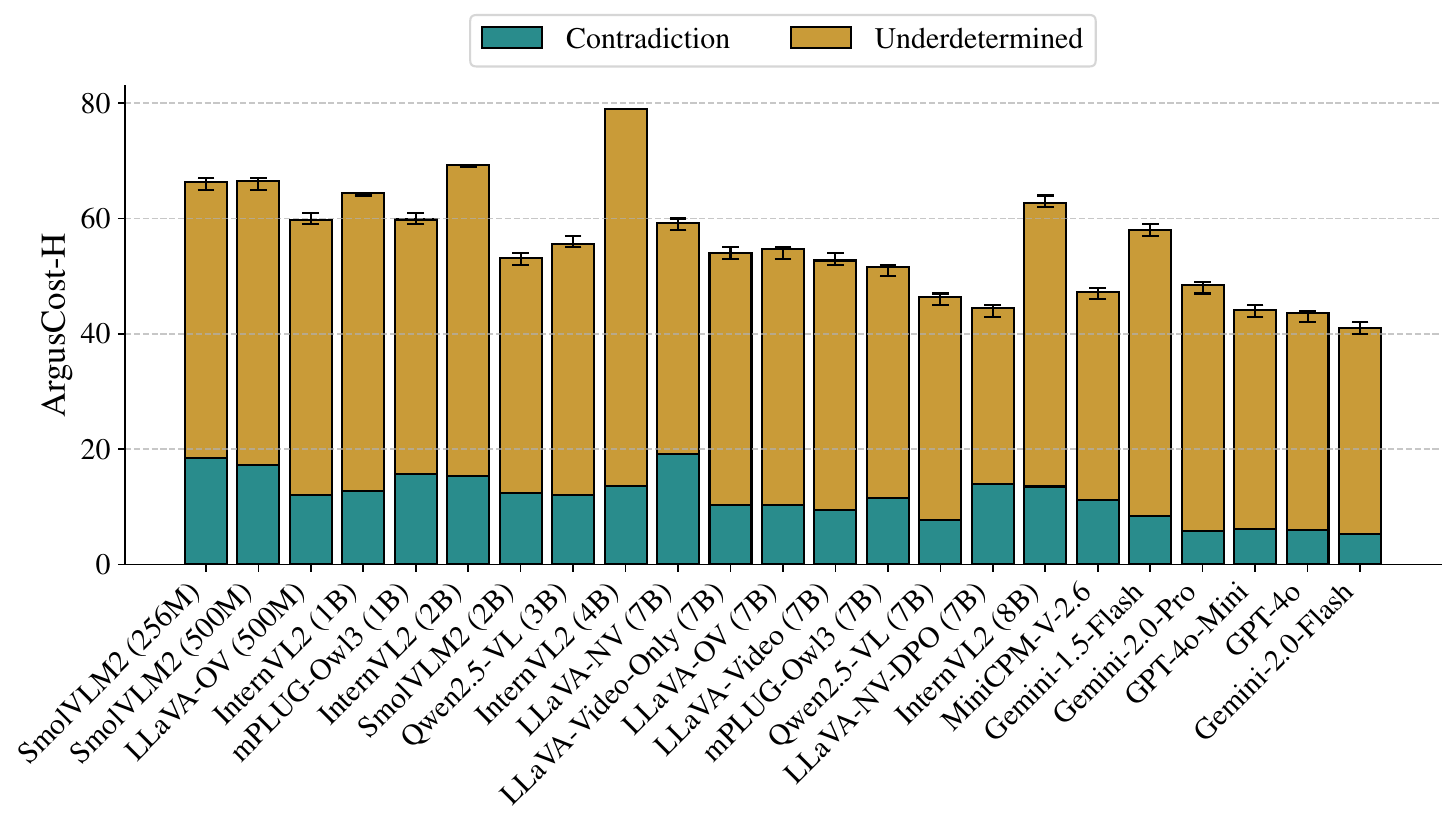}
    \caption{Breakdown of hallucination errors into ``contradiction" and ``undetermined". Undetermined errors dominate the majority of errors, across model families and sizes.}
    \label{fig:hallucination_type}
\end{figure}

In Section~\ref{sec:results}, we analyzed the \hscoreh breakdown by sentence type (summary, visual description, dynamic action). Another way to examine \hscoreh is by categorizing hallucination errors as either ``contradiction" or ``undetermined." Figure~\ref{fig:hallucination_type} visualizes this breakdown, showing that approximately two-thirds of errors fall into the "undetermined" category, while the remaining one-third are contradictions. This trend holds across all models and sizes, indicating that models generate more fabricated content rather than directly contradicting the input. Additionally, as models improve (moving right on the x-axis), the proportion of undetermined errors increases, suggesting that improvements primarily come from reducing contradiction errors, leading to a better overall understanding of the video. This observation aligns with the correlation between \hscoreh and \hscoreo, as models with fewer contradictions also tend to omit less information.

\paragraph{Cost-Breakdown by Cost-Type}

\begin{figure}
    \centering
    \includegraphics[width=\linewidth]{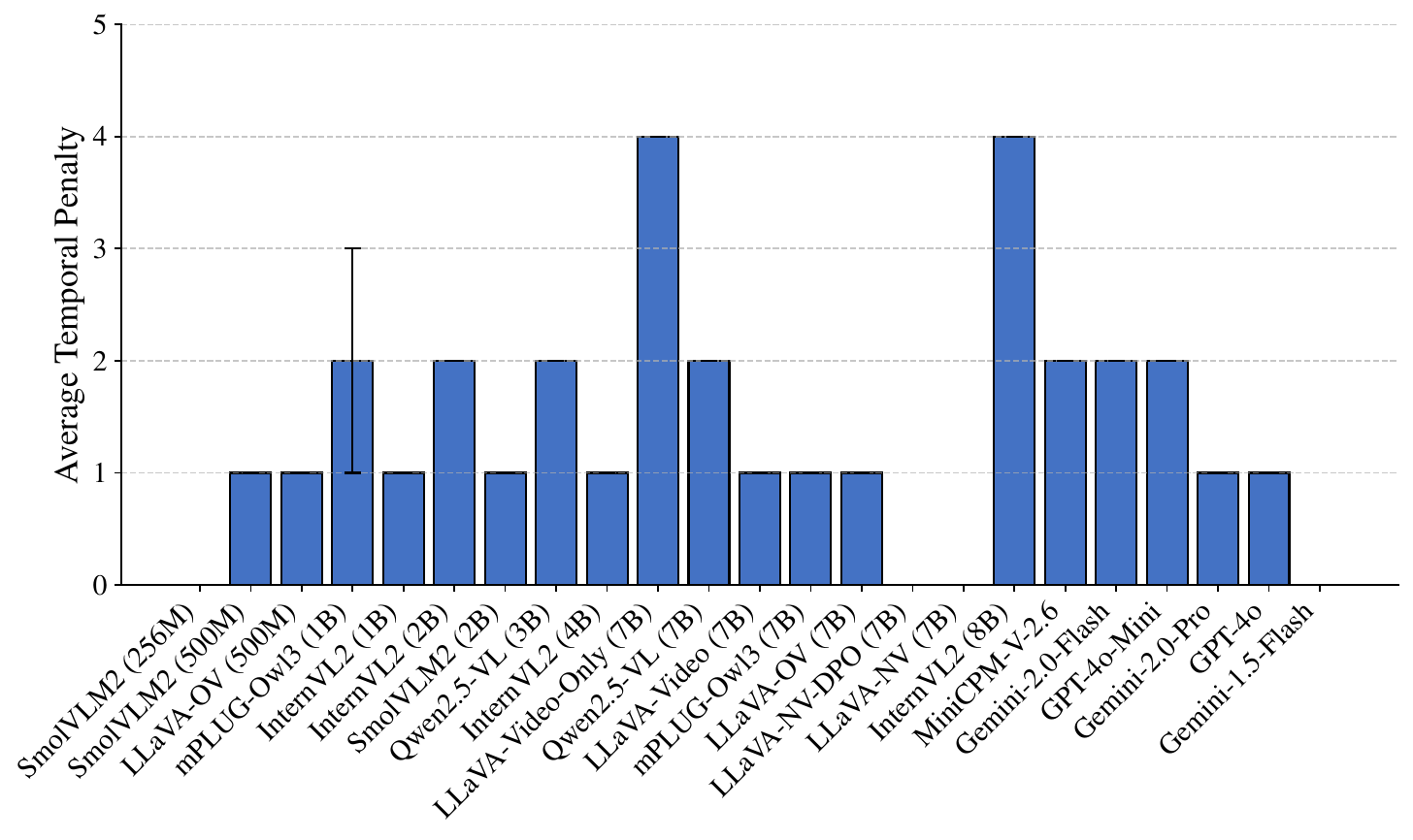}
    \caption{Contribution of Temporal Penalties in the overall cost. Temporal penalties make a minor proportion of the total costs.}
    \label{fig:temporal_penalty}
\end{figure}

Another way we break down \hscoreh is by examining the Base Costs and Temporal Penalties defined in Section~\ref{sec:methods}. In Figure~\ref{fig:temporal_penalty}, we visualize the temporal penalties incurred by different models. Since these penalties are much smaller than the base cost, making up only about 1\% to 4\% of the total cost, we present the temporal penalty costs seperately. We observe that for most models, temporal penalties are close to zero, with a maximum of 4\% for LLaVa-Video and InternVL2 (8B). Our qualitative analysis suggests that weaker models often fabricate information or produce contradictions, meaning they fail to generate dynamic actions in the first place, resulting in minimal temporal penalties. In contrast, stronger models still make some errors but tend to preserve the correct event order for the actions they get right, leading to lower temporal penalties overall.

\begin{figure}
    \centering
    \includegraphics[width=0.5\linewidth]{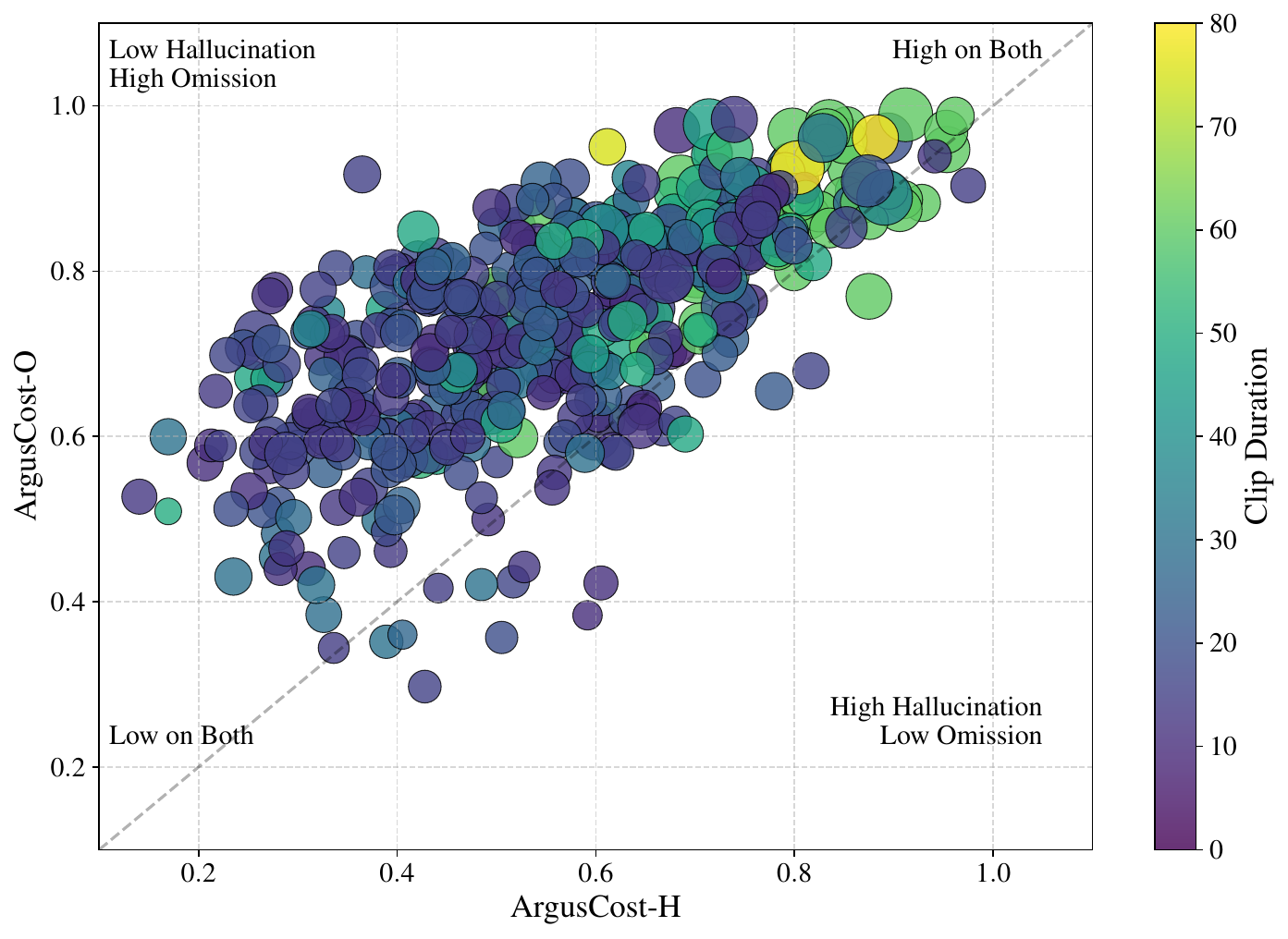}
    \caption{Visualizing the correlation between clip duration, \hscoreh and \hscoreo, and caption density}
    \label{fig:clip_duration_eff}
\end{figure}

\paragraph{Effect of Clip Duration}
In Section~\ref{sec:results}, we discussed the mild positive correlation between \hscoreh, \hscoreo, and clip duration. Figure~\ref{fig:clip_duration_eff} provides a visualization to further analyze this effect. The figure plots \hscoreh and \hscoreo, with clip duration represented by colorbar hue and model-generated caption size indicated by marker size. The correlation is evident as brighter markers (longer videos) are clustered in high hallucination and omission areas, while darker markers are positioned elsewhere. Additionally, marker size is skewed toward the top-right, indicating that longer videos tend to generate denser captions, as expected.

\begin{figure}
    \centering
    \includegraphics[width=\linewidth]{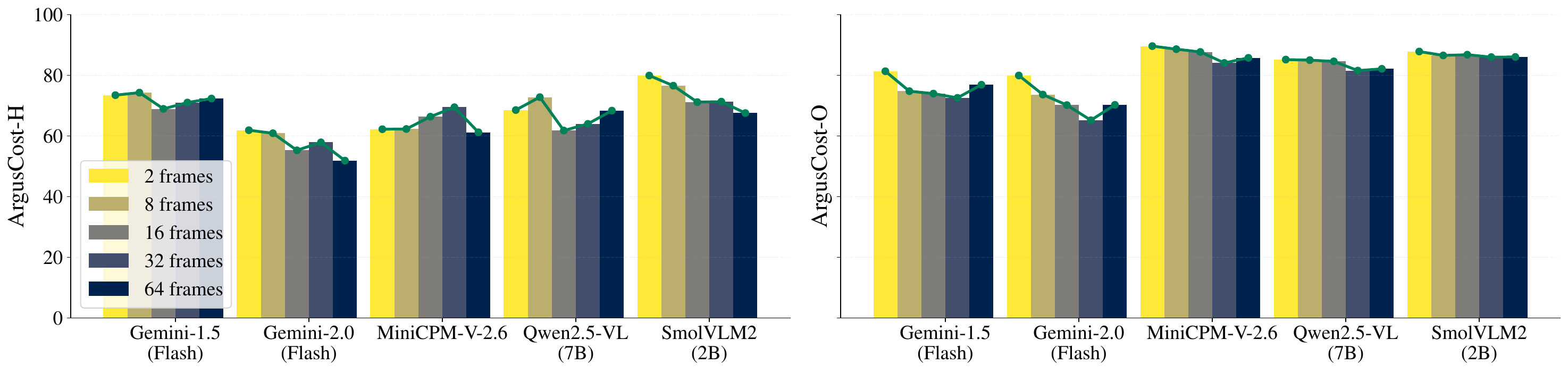}
    \caption{Effect of total number of frames on \hscoreh and \hscoreo. Gemini models and SmolVLM2 show consistent reduction in hallucinations with more frames, while MiniCPM-V and Qwen2.5-VL exhibit fluctuating levels of hallucinations (as measured by \hscoreh). All models show consistent improvement in \hscoreo results as we increase the number of frames.}
    \label{fig:eff_frame_rate_omm}
\end{figure}

\paragraph{Effect of Frames on Omission}
We discussed the effect of increasing the total number of frames provided to model from 2 to 64 on \hscoreh in the main paper. Here, in Figure~\ref{fig:eff_frame_rate_omm} we discuss the effect on \hscoreo. We note that omission rate consistently decreases as the number of frames increase for all the models. Specifically, when moving from 2 to 64 frames, Gemini-1.5-Flash improves by 4.45\%, Gemini-2.0-Flash improves by 9.69\%, MiniCPM-V-2.6 improves by 3.86\%, Qwen2.5-VL improves by 3.04\%, SmolVLM2 improves by 1.78\%.

\section{Relationship With Video Characteristics}
\label{appendix_sec:rel_video_characteristics}

In the main paper, we discussed the correlation between various video characteristics—such as video duration—with both \hscoreh and \hscoreo. In this section, we expand on that analysis by examining a broader range of video characteristics and their specific relationship with \hscoreh, aiming to provide a more comprehensive understanding of the influence of video content.

\begin{enumerate}
    \item \textbf{LAION Aesthetic Score} \cite{schuhmann2022laion_aesthetics_predictor_v1}: The LAION aesthetic score is commonly used to assess the aesthetic quality of an image (ranging between 1-10). In our setting, we adapt this score by computing the average aesthetic score across all frames in a video. While this approximation has limitations, it serves as a reasonable starting point. One hypothesis is that certain models—depending on their training—may perform worse on videos with lower aesthetic quality, or conversely, may be biased towards more aesthetically pleasing content.

    \item \textbf{Lighting Unique Count}: We utilize a shot detection model i.e. \texttt{diffusers/shot-categorizer-v0} \cite{huggingface2025shot_categorizer_v0} to identify the lighting type in each frame of a video. We then compute the number of unique lighting types (e.g., Daylight, Sunny) and use this count as a metric. The underlying assumption is that videos with rapid or frequent lighting changes may pose challenges for models, potentially increasing the likelihood of hallucinations.

    \item \textbf{Subject Presence}: The number of distinct entities present in a video may also influence model performance, with more entities potentially leading to increased confusion for VideoLLMs. To quantify this, we employ the Segment Anything Model (SAM) \cite{ravi2024sam} to generate segmentation masks and count the number of unique entities per video. We then analyze the distribution of these counts across the dataset.
\end{enumerate}

\begin{figure}
    \centering
    \includegraphics[width=0.8\linewidth]{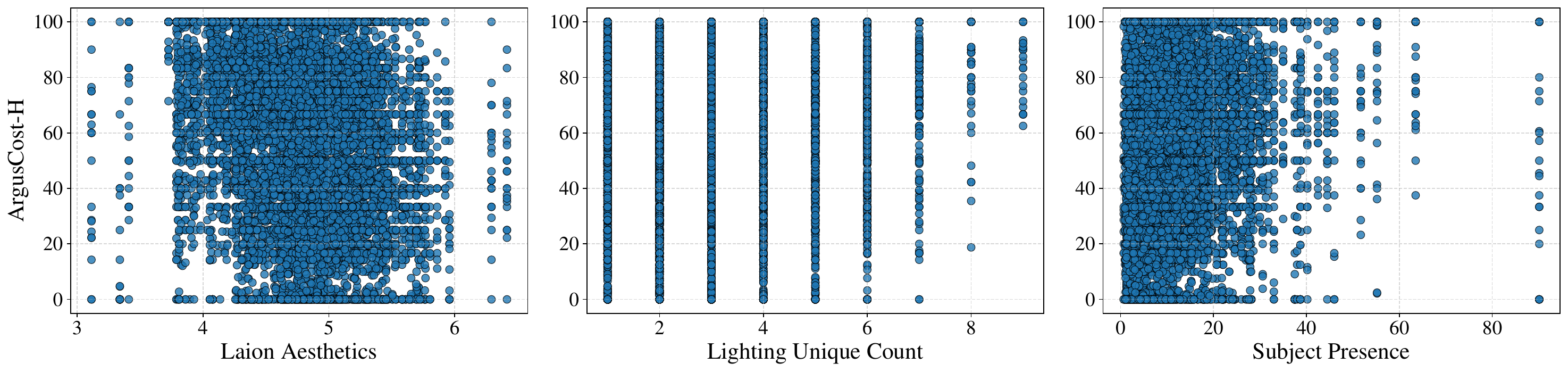}
    \caption{Correlation between various video-characteristics and \hscoreh.}
    \label{fig:agg_filter_results}
\end{figure}

\begin{figure}
    \centering
    \includegraphics[width=0.8\linewidth]{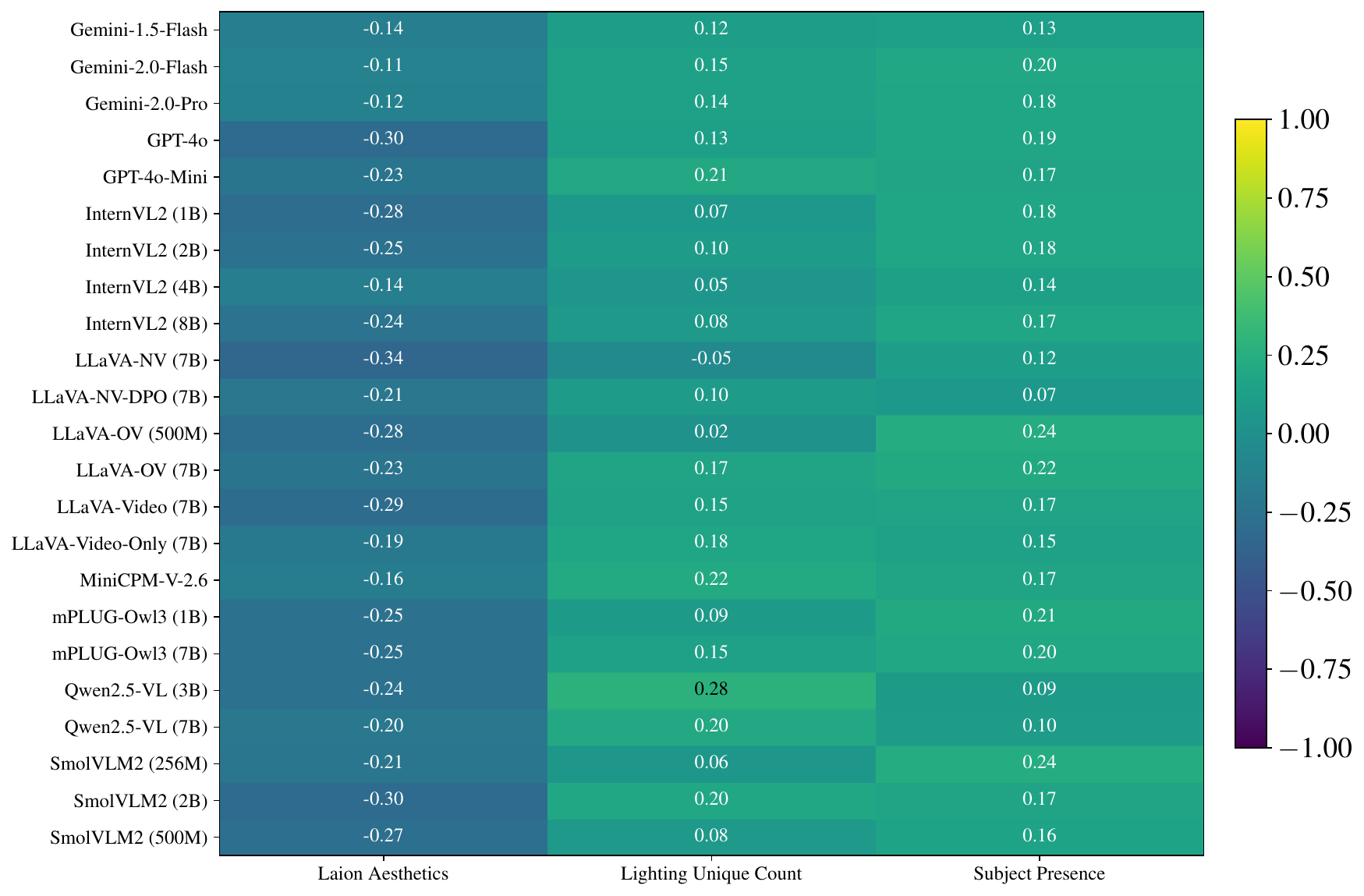}
    \caption{Model-wise correlation heatmaps between video filter metrics and \hscoreh.}
    \label{fig:model_wise_filt_results}
\end{figure}

We visualize these results in \cref{fig:agg_filter_results}. For the metrics examined, we do not observe a strong or consistent correlation with \hscoreh. It is important to note that this is an initial analysis and not intended to be exhaustive. In future work, we plan to explore more comprehensive features, including those that capture motion dynamics and other relevant characteristics. In \cref{fig:model_wise_filt_results}, we visualize the correlation heatmaps separately for each model. Among the filters, we observe that the models exhibit the strongest (negative) correlation with the LAION aesthetic score.

\section{Additional LLM-as-a-judge Details}
\label{appendix_sec:llm_entailment_judge_details}

\subsection{Extended Discussion on Type and Verdict Categorization}
\paragraph{Type Categorization. } We observed that Video-LLMs primarily generate three types of sentences: (1) summary sentences that provide a high-level overview, often at the start of a caption; (2) visual descriptions that highlight key attributes of entities in the video; and (3) dynamic action descriptions that capture important events. Categorizing sentences by type allows us to analyze which kinds of hallucinations these models tend to produce.  
Additionally, dynamic actions have an inherent temporal structure, meaning the order in which they appear is crucial. To account for this, we introduce an order penalty when the model-generated sequence deviates from the source descriptions. However, summary and visual description sentences (assuming the visual attributes remain unchanged) do not follow a strict temporal order, making their placement in the generated captions more flexible. 
By labeling each sentence type, we ensure that only dynamic actions are evaluated for order discrepancies. Intuitively, our temporal penalty operates on the principle that dynamic-action sentences in the VideoLLM caption should follow the same order as their evidence lines in the human caption. Thus, by aligning generated dynamic actions with their corresponding source events, we can systematically compare event sequences and quantify any inconsistencies in temporal order.

\paragraph{Verdict Categorization. } Our classification approach is grounded in the natural language inference (NLI) framework, which defines three key relationships between a generated sentence and the source descriptions: entailment (clear supporting evidence), contradiction (clear evidence against), and undetermined (insufficient evidence to confirm or refute). Contradictions represent a specific type of hallucination where the model generates content that directly conflicts with the source. For example, if a video shows a man cooking in a kitchen but the model, due to pretraining biases, describes a woman cooking, this constitutes a contradiction. On the other hand, undetermined cases arise when the model includes details that cannot be confirmed or denied based on the available evidence. For instance, if a blurred shot in the video shows a person cooking and the human-annotated caption states ``a person is cooking," but the model-generated caption specifies ``a woman is cooking," the gender detail is undetermined. While both contradiction and undetermined cases are forms of hallucination, they differ in nature. Distinguishing between them allows us to categorize errors more precisely and measure their effects separately, leading to a more nuanced evaluation of model reliability.

\subsection{Computational Costs}

\paragraph{Monetary Costs for NLI Evaluation using GPT-4o:}
\begin{itemize}
    \item Input Tokens: 843 (Base Prompt Instructions) + 477 (Average Human Caption Tokens) + 234 (Average Model-Generated Tokens) + 910 (In-Context Examples) = 2464
    \item Cost per Input Token: \$2.50 per 1M tokens
    \item Input Cost per Video: $\$0.00616$
    \item Average Output Tokens: 640
    \item Cost Per Output Tokens: \$10.0 per 1M tokens
    \item Output Cost per Video: 0.006
    \item Cost Per Video: $\$0.01216$
\end{itemize}

This is an upper-bound on the total cost, since the base instruction tokens can be cached. Additionally, as we see that SOTA open source also correlate highly with GPT-4o's judgment, so one could potentially use them as well.

\subsection{Prompt}
\begin{verbatim}
You are given two inputs: a source_caption (serving as the premise) 
and a target_caption (serving as
a  series of hypothesis lines). For each line in the target_caption, evaluate 
its relationship to 
the source_caption according to the following guidelines:

- **Entailment:**  The target line is fully supported by the source_caption. 
This includes cases where the target uses alternative phrasing or 
synonymous descriptions, and minor 
variations in attributes (for example, "pink" (or "red") and "reddish pink" are treated 
as equivalent). Additionally, If the target line asserts that an aspect 
(such as lighting, background consistency, or non-occurrence of an event) 
remained unchanged or did not happen, and the **source_caption** 
is silent on that aspect, treat it as entailment since if there were a change the 
**source_caption** 
would have mentioned it.  
Similarly, also include natural or highly probable attributes/states
that would only be mentioned in the source if they deviated from the norm. 
For example: If there's mention of an airplane, 
we can assume it's large, as most 
airplanes naturally are. However, this only applies to immediately obvious 
and universally expected attributes. Any substantial elaboration or 
specific details 
beyond the most basic expectations should still be treated as undetermined.

- **Contradiction:**  The target line contains a direct conflict with the 
information  in the source_caption. This means that the target asserts a fact 
or detail that explicitly opposes a fact stated in the source_caption 
(for example,  attributing an action to Person-X in the source versus 
Person-Z in the target). 

- **Undetermined:**  The target line introduces additional details or makes occurrence 
or attribute claims that are semantically independent 
from the information provided in the 
source_caption. In these cases, the source_caption neither provides strong evidence to 
support the extra details nor directly contradicts them. This category also 
covers instances where coreference between events or entities is ambiguous—for example, 
when it is unclear whether a new event or entity mentioned in the target corresponds 
to one in the source. In such cases, because the evidence is insufficient 
to clearly support or refute the claim, the relationship 
should be classified as undetermined.

For each line in the **target_caption**, first output the information category, 
i.e., either it's a summary sentence (summary), or 
describing a static visual detail of the
video like color of an entity (visual-description), or a dynamic action that 
includes events, 
attribute and relationship changes, etc (dynamic-action).  Next output an evidence line 
or phrase 
from the **source_caption** that serves as the basis for your verdict. If no evidence
exists, 
use an empty string. Then, provide reasoning for your verdict based on the evidence, 
followed by the final classification: "entailment," "contradiction," 
or "undetermined." Do not include any additional commentary or summary.

**Output Format:**  
Line {num}: {Line text}
    - Type: {summary/visual-description/dynamic-action}
    - Evidence: {quote a specific line/phrase from the source upon 
    which to base your verdict; use an empty string if no evidence exists.}  
    - Reasoning: {explanation for the verdict based on the evidence}  
    - Verdict: {entailment/contradiction/undetermined}


**Final-Instructions:**

1. Treat the **source_caption** 
as the premise and each line of the **target_caption** as a 
hypothesis.
2. For each hypothesis line, decide if it is:

   - **Entailment**: Completely supported by the source_caption.
   
   - **Contradiction**: Contains information that 
   directly conflicts and is unsupported by 
   the source_caption.
   - **Undetermined**: Contains additional details or
   claims that the source_caption neither 
   strongly supports nor directly contradicts. Mostly 
   includes cases of ambiguous coreference 
   between entities or events.
3. Follow the output format exactly as specified, without
any extra summaries or notes.

---

In-Context Examples:

{IN_CONTEXT_EXAMPLES}

---

**source_caption:**
"""
{source_caption}
"""

**target_caption:**
"""
{target_caption}
"""
\end{verbatim}
\section{Details Regarding Qualitative Examples}
\label{appendix_sec:details_qual_examples}

\subsection{Figure-1 Details.}

\textbf{Human Caption:} \begin{verbatim}
The black title screen reads “SUNFEAST PASTA TREAT, cooking, 45 sec., 14th May 2008” 
in white font. A boy enters his home holding a football and wearing a green t-shirt.
A woman, wearing a pink top, calls him over to the kitchen and asks him to explain
how “Sunfeast Pasta” is cooked. The boy stands on a stool and puts on a white chef’s
hat. He jumps off the stool and seems to signal to the woman to get something. Now,
the boy is holding a pair of scissors and cuts the pasta packet where the brand name
"Sunfeast" is clearly visible. The woman shows the boy some teacups. We see a shot of
boiling water. The woman passes the pasta to the boy, followed by a shot of the pasta
being added to the boiling water. We see spices being added to the pasta. The boy
grabs a cooking stick and stirs the pasta. The woman puts the garnished pasta on a
plate, while the boy places a white cloth on the table. The woman appears impressed
and places the pasta on the table. Then we see a paper with the pasta recipe: 2 tea
cups of water, pasta, sauce mix, and 8 minutes. The woman eats the pasta and looks
impressed. She salutes the boy, who bows. At the end, we see the company’s brand
name, logo, their tagline "Quick and easy".
\end{verbatim}

\textbf{Gemini-Flash-2.0 Caption:} \begin{verbatim}
Here's a detailed description of the video, based on the images provided:

**General Impression:**

The video appears to be a commercial for Sunfeast Pasta Treat, a quick-and-easy pasta
product.  It features a mother and son preparing the pasta together. 
The overall tone is upbeat and suggests convenience.

**Key Visual Elements:**

1.  **Introductory text on a black screen:** The video starts with text stating
"SUNFEAST PASTA TREAT," followed by "*cooking*," "45 sec." and the date "14th May, 2008."
2.  **Family Scene:** The initial scenes show a child wearing a green shirt and a chef's hat 
looking at his mother standing near a kitchen shelf
3.  **Product Focus:** The video clearly highlights the packaging of the "Sunfeast
Pasta Treat" product, with the bag appearing close up to the camera. A pair of hands
is shown cutting open the product's packaging. 
The contents of the pasta treat are displayed.
4.  **Cooking Process:** We see shots of the pasta being cooked, likely to emphasize
the simplicity of the preparation.
5.  **Appearance of the finished product:** The finished pasta is served on a white
plate, decorated with herbs.
6.  **Mother in the car:** There are multiple shots of a woman in the car.
7.  **Child's "Recipe Card":** Briefly, there is a close-up of a handwritten note 
that contains instructions.
8.  **"Quick and Easy" Text Overlay:** Text emphasizing the convenience of the 
product appears on screen, accompanying a display of the pasta varieties.

**Overall Narrative:**

The commercial likely follows this narrative:

*   Introduction: Setting the scene, indicating it's a cooking segment/advertisement.
*   Family Cooking: It looks to feature the child in the kitchen with his mother.
*   Showcasing product: Emphasizing the convenience and ease of use.
*   End: the "Quick and Easy" tagline to reinforce the product's core benefit.
\end{verbatim}

\subsection{Figure-2 Details.}

\textbf{Human Caption:} \begin{verbatim}
In a natural daylight setting, the video begins with a close-up shot of a thin,
grayish tree branch against a blurred background of green foliage. The branch splits
into two sub-branches in the middle and extends slightly upward toward the left side
of the screen.
On the rightmost edge of the frame, the chin and nose of an animal become visible. It
appears reddish-brown, with a sharply triangular-shaped snout, while the rest of its
body remains unseen. As the video progresses, the animal moves from right to left,
gradually revealing its large, expressive eyes that move slowly and deliberately.
Around its nose and eyes, intricate dot-like patterns emerge, varying in color—some
blending with its skin while others, particularly between the eye and stomach, take
on a grayish hue. These grayish patterns extend beyond just dots, forming irregular
patches along the face and stomach area. Below its chin, faint grill-like structures
can also be seen.
The animal is a chameleon. As it moves forward, it lifts its left limb and grips the
portion of the branch ahead. After this initial movement, it briefly pauses, its ever
-shifting eyes meticulously scanning the surroundings. Then, in a sudden
transformation, its reddish skin changes to a bright pink, though the gray dots and
patches remain intact. After the color transformation, the chameleon continues
forward, now using its right limb to grasp the portion of the branch ahead,
momentarily flicking out its tongue while its eyes persist in their careful
assessment of the environment.
Following another brief pause, the chameleon reaches the point where the branch
splits, its mouth now close to the division. In an instant, its color shifts once mor
e—from bright pink to green—while the gray dots remain unchanged. During this
transition, an intermediate blend of orange and yellow briefly appears. The chameleon
then holds its position, its eyes still in constant motion, meticulously observing
its surroundings. The background remains consistently blurred throughout. 
\end{verbatim}

\textbf{LLaVa-OV-7B:} \begin{verbatim}
    The video begins with a close-up of a thin, gray branch set
    against a blurred green background, likely foliage. A red
    chameleon with distinctive patterns on its skin appears and
    starts to climb the branch. The chameleon's eyes are large
    and expressive, and it moves slowly and deliberately,
    showcasing its unique features. The scene transitions to
    another red chameleon with similar patterns, also climbing
    the branch. This chameleon is more vibrant in color and has a
    slightly different pattern compared to the first one. The
    background remains blurred green foliage, emphasizing the
    chameleon's movement and interaction with the branch. The
    video then shifts to a green chameleon with white spots on
    its skin, also climbing the same thin, gray branch. The green
    chameleon's eyes are large and expressive, and it moves
    slowly and deliberately, showcasing its unique features. The
    background remains blurred green foliage, emphasizing the
    chameleon's movement and interaction with the branch. The
    video concludes with the green chameleon continuing to climb
    the branch, highlighting its vibrant color and distinctive
    patterns.
\end{verbatim}

\end{document}